\documentclass[10pt,twocolumn,letterpaper]{article}

\usepackage[pagenumbers]{cvpr} 

\usepackage[dvipsnames]{xcolor}

\usepackage{graphicx}
\usepackage{epigraph,varwidth}
\usepackage{booktabs}   
\usepackage{multirow}   
\usepackage{array}      
\usepackage{colortbl} 
\usepackage{tabularx} 
\usepackage{graphicx} 
\usepackage{booktabs} 
\usepackage{caption} 

\definecolor{cvprblue}{rgb}{0.21,0.49,0.74}
\usepackage[pagebackref,breaklinks,colorlinks,citecolor=cvprblue]{hyperref}

\def\paperID{XXXX} 
\def\confName{CVPR}
\def\confYear{2025}

\title{Probing the Mid-level Vision Capabilities of Self-Supervised Learning}

\author{
    Xuweiyi Chen$^{1}$ \quad
    Markus Marks$^{2}$ \quad
    Zezhou Cheng$^{1}$ \\
    $^{1}$University of Virginia \quad $^{2}$California Institute of Technology \\
    \url{https://midvision-probe.cs.virginia.edu/}
}

\begin{document}
\maketitle
\begin{abstract}

Mid-level vision capabilities --- such as generic object localization and 3D geometric understanding --- are not only fundamental to human vision but are also crucial for many real-world applications of computer vision.
These abilities emerge with minimal supervision during the early stages of human visual development. 
Despite their significance, current self-supervised learning (SSL) approaches are primarily designed and evaluated for high-level recognition tasks, leaving their mid-level vision capabilities largely unexamined. 

In this study, we introduce a suite of benchmark protocols to systematically assess mid-level vision capabilities and present a comprehensive, controlled evaluation of 22 prominent SSL models across 8 mid-level vision tasks.
Our experiments reveal a weak correlation between mid-level and high-level task performance. 
We also identify several SSL methods with highly imbalanced performance across mid-level and high-level capabilities, as well as some that excel in both. 
Additionally, we investigate key factors contributing to mid-level vision performance, such as pretraining objectives and network architectures.
Our study provides a holistic and timely view of what SSL models have learned, complementing existing research that primarily focuses on high-level vision tasks.
We hope our findings guide future SSL research to benchmark models not only on high-level vision tasks but on mid-level as well.

\vspace{-10px}
\end{abstract} 
\section{Introduction}
\label{sec:intro}

\begin{figure}[ht!]
    \centering
    \includegraphics[width=\linewidth]{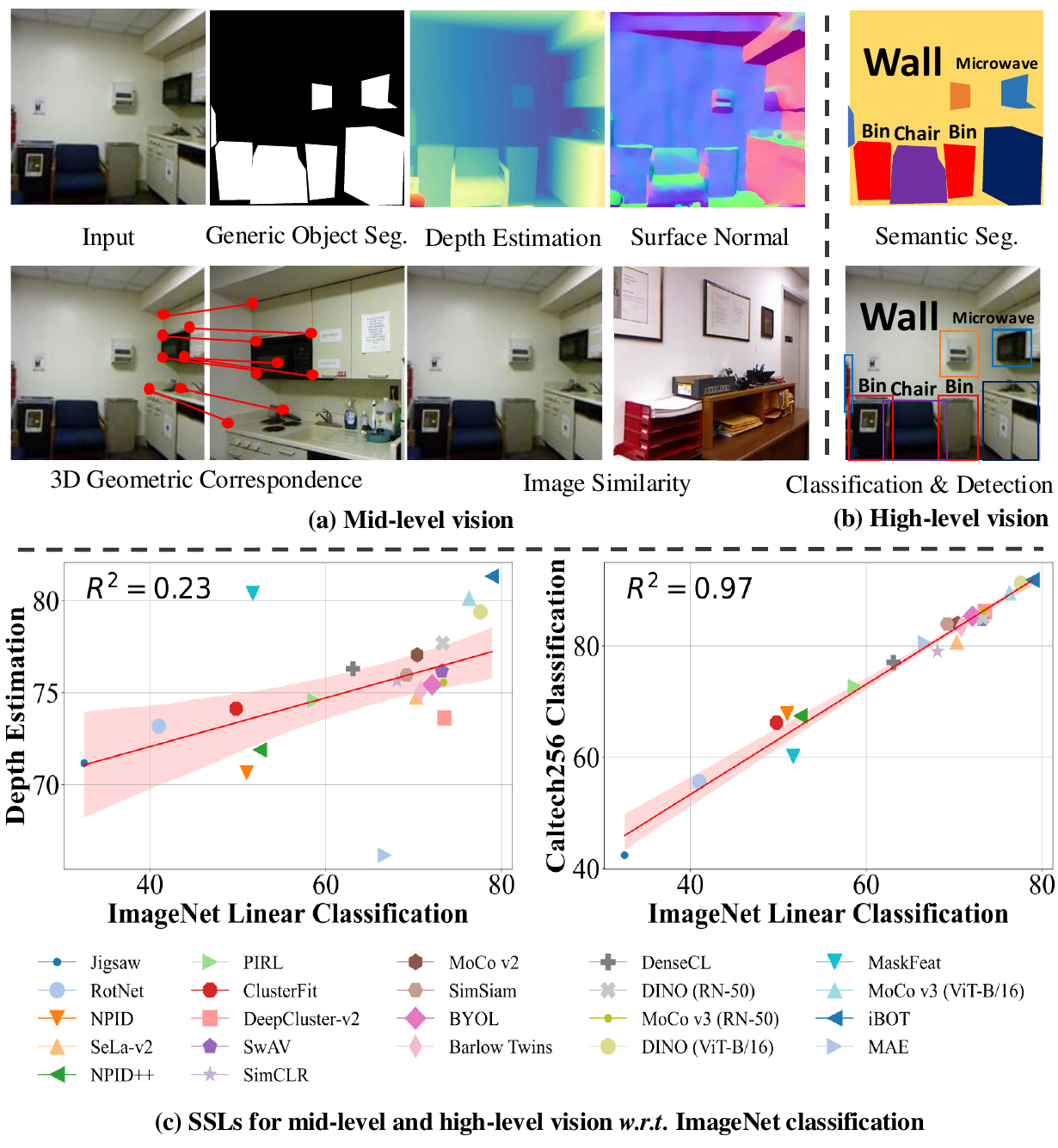}
    \vspace{-4mm}
    \caption{\textbf{Mid-Level Vision vs. High-Level Vision Tasks.} 
    We provide a comprehensive evaluation of prominent self-supervised learning methods (SSLs) across a wide range of mid-level vision tasks \textbf{(a)}, complementing the standard evaluation in high-level vision tasks \textbf{(b)}. Although SSL performance in mid-level vision tasks (\eg, depth estimation) is positively correlated with ImageNet linear probing \textbf{(c, left)}, this correlation is much weaker than that observed among high-level vision tasks (\eg, Caltech-256~\cite{griffin2007caltech} vs. ImageNet classification) \textbf{(c, right)}, as indicated by the $R^2$ statistics. \vspace{-10pt}
    }
    \label{fig:teaser}
\end{figure}

Mid-level vision capabilities~\cite{marr2010vision} (\cref{fig:teaser}a) play a vital role in human visual development and daily life activities.
By the age of one, children can re-organize retinal images into objects (\ie, \emph{perceptual grouping}), construct the visual world in three dimensions (\ie, \emph{3D geometric understanding}), track moving objects (\ie, \emph{cross-view correspondence}), grasp and bite these objects without knowing their semantics, and purposefully navigate through the 3D world.
Importantly, such mid-level perception is developed without supervision~\cite{hoffman2000visual}.

On the other hand, self-supervised learning (SSL) in computer vision has made significant strides~\cite{Noroozi2016,Gidaris2018,Wu2018a,misra2019pirl,chen2020simple,chen2020improved,chen2021exploring,caron2021emerging,oquab2023dinov2,zhou2021ibot}. The goal of SSL is to learn visual representations from unlabeled data so that these pretrained representations can be applied to a wide range of downstream tasks, reducing the need for human annotations and enhancing learning efficiency and performance.
However, SSL is predominantly evaluated based on performance in high-level visual recognition tasks, such as image classification, object detection, and semantic segmentation (\cref{fig:teaser}b). At the same time, the mid-level vision capabilities of SSL models are largely under-explored.
To fill this gap, this paper addresses the following questions:

\begin{itemize}
    \item \emph{Does strong high-level vision performance of SSLs indicate strong mid-level vision capabilities?}
    \item \emph{What makes an SSL model effective for mid-level vision tasks?}
\end{itemize}

To answer these questions, we comprehensively evaluate SSLs in a wide range of mid-level vision tasks, including generic object segmentation, monocular depth and surface normal estimation, geometric correspondence, and mid-level image similarity (Fig.~\ref{fig:teaser}a and Tab.~\ref{tab:tasks}).
We then relate the mid-level vision performance of these SSLs to their high-level vision performance.
Our evaluations covers 22 seminar SSL approaches (Tab.~\ref{tab:ssls}) from diverse categories developed in the past decades (\eg pretext tasks~\cite{Noroozi2016,Gidaris2018}, contrastive learning~\cite{Wu2018a,chen2020simple,chen2021exploring,grill2020bootstraplatentnewapproach,chen2021empiricalstudytrainingselfsupervised}, self-distillation~\cite{caron2021emerging}, and generative modeling~\cite{he2021maskedautoencodersscalablevision, wei2022masked}). 
To make a controlled study and fair comparisons across different SSLs, 
our study primarily focuses on SSLs with publicly available checkpoints pretrained on ImageNet1K~\cite{deng2009imagenet}. 
This allows us to gain insights into what factors may contribute to the mid-level vision capabilities in SSL (\eg, training objectives, network architectures, \etc).

Our observations are as follows:
\begin{itemize}[itemsep=1pt, topsep=1pt]
\item The mid-level vision capabilities of SSLs are positively correlated with high-level vision. However, that correlation is generally weaker than that among high-level vision tasks (Fig.~\ref{fig:teaser}c and Fig.~\ref{fig:mid-vs-imagenet}).
\item Some SSLs have highly imbalanced mid-level and high-level vision capabilities. 
For instance, MAE~\cite{he2022masked} under-performs most SSLs in the mid-level vision tasks while having competitive high-level vision performance; MaskFeat~\cite{wei2022masked} has better mid-level performance than most SSLs with similar high-level vision (Fig.~\ref{fig:ranking}).
\item Network architectures, model capacity, and pretraining strategies impact mid-level vision capabilities. In general, ViT~\cite{dosovitskiy2020image} outperforms ResNet~\cite{he2016deep};  
Increasing the network capacity is beneficial (Tab.~\ref{tab:architecture}); 
The leading methods are generally based on generative modeling (Fig.~\ref{fig:ranking}).
\end{itemize}

Our study is built upon Probe3D~\cite{el2024probing}, a recent work that evaluates the 3D awareness of visual models. In this work, we explore a broader range of mid-level vision tasks.
Furthermore, our study primarily focuses on SSLs pretrained exclusively on ImageNet1K, while Probe3D probes visual foundation models that are trained from very diverse training configurations such as varying supervision signals (\eg CLIP~\cite{radford2021learning}) and dataset sizes. 
Our controlled settings allow us to gain insights into the factors contributing to model performance.
Our exploration is also closely related to prior works that benchmark SSLs in various vision tasks~\cite{Ericsson2021HowTransfer,goyal2019scaling,Newell_2020_CVPR,tong2024cambrian}, such as out-of-domain classification~\cite{marks2024closer} and fine-grained recognition~\cite{van2021benchmarking,su2021realistic}.
Unlike these works, our evaluation includes more recent SSLs, providing a timely study on this topic.

In summary, our study offers a comprehensive view of the representations learned by SSL models. 
We hope these findings inspire practical approaches that perform well across both mid-level and high-level vision tasks. We will release our benchmark suite to support future SSL benchmarking in the research community.

\section{Related Work}
\label{sec:related}
\paragraph{\textbf{Self-Supervised Learning (SSL).}}
The goal of self-supervised learning (SSL) in computer vision is to pretrain image representations using unlabeled data, enabling these representations to be transferred effectively to various downstream tasks. Early SSL methods leveraged a range of pretext tasks, such as colorization~\cite{Zhang2016}, rotation prediction~\cite{Gidaris2018}, jigsaw puzzle solving~\cite{Noroozi2016}, and inpainting~\cite{Pathak2016}. 
Recent advances have focused on contrastive learning~\cite{Wu2018a,chen2020simple,he2019momentum, chen2021exploring, grill2020bootstraplatentnewapproach, zbontar2021barlowtwinsselfsupervisedlearning} and generative modeling~\cite{he2022masked, wei2022masked, peng2022beitv2maskedimage, liu2023pixmimrethinkingpixelreconstruction, dong2023maskclipmaskedselfdistillationadvances}.
Despite extensive research, SSL performance is primarily evaluated on high-level vision tasks (\eg, ImageNet classification, object detection), overlooking the practical relevance of mid-level vision tasks. In this work, we focus on evaluating SSL methods on mid-level vision tasks.
\vspace{-10pt}
\paragraph{\textbf{Self-Supervised Pretraining for 3D Vision.}}
Unlike most SSL methods focusing on 2D visual recognition tasks, a few recent works explore SSLs for 3D vision. 
For example, CroCo~\cite{weinzaepfel2022croco} adopts masked image modeling to learn 3D representations from image pairs showing the same scene from different viewpoints. 
Despite the improvement in downstream 3D vision tasks, 
CroCo compromises high-level recognition performance.
As a followup, CroCo-v2~\cite{croco_v2} extends the training data of CroCo to a large dataset of natural images and demonstrates improved performance in downstream optical flow and stereo vision tasks.
However, learning representations for 3D vision tasks remains under-explored. 
\vspace{-10pt}
\paragraph{\textbf{Benchmarking Self-Supervised Learning.}} 
Numerous studies underscore the growing importance of empirically evaluating general-purpose representation learning. For example, Goyal~\etal~\cite{goyal2019scaling} provide extensive benchmarking of SSLs across multiple downstream tasks, including classification, navigation, object detection, and surface normal estimation. Newell~\etal~\cite{Newell_2020_CVPR} examine the effectiveness of SSLs under various configurations (\eg, annotation quantities) in downstream tasks such as pose estimation and depth estimation. Ericsson~\etal~\cite{Ericsson2021HowTransfer} evaluate 13 SSLs across 40 downstream vision tasks, demonstrating a weak correlation between SSL performance on ImageNet classification and tasks like few-shot learning, object detection, and dense prediction. More recently, Marks~\etal~\cite{marks2024closer} studied how classification-based SSL evaluation protocols correlate with and predict downstream performance on out-of-domain datasets.
In line with these works, we comprehensively evaluate existing SSLs across various downstream tasks. 
We focus on mid-level vision performance and how it relates to high-level vision.

\paragraph{\textbf{Benchmarking Visual Foundation Models.}} 
More recently, several studies have evaluated large visual foundation models (VFMs) that are not exclusively trained by self-supervised learning methods (\eg, CLIP~\cite{radford2021learning}). For example, Probe3D~\cite{el2024probing} examines the 3D awareness of VFMs~\cite{caron2021emerging,kirillov2023segment}, while Cambrian-1~\cite{tong2024cambrian} utilizes large language models (LLMs) and visual instruction tuning to assess visual representations across various 2D and 3D vision tasks. Since VFMs are typically trained with diverse approaches (\eg, different supervision signals, datasets, and architectures), isolating the factors contributing to the performance of the representations they learn is challenging.
In contrast, we conduct a controlled evaluation of SSL models that provide publicly available checkpoints pretrained on ImageNet1K. 
This approach enables fair comparisons across SSLs and offers deeper insights into their methodologies.
\section{Background: Vision Task Taxonomy}

There exist numerous computer vision tasks~\cite{taskonomy2018} which can be categorized from different perspectives. 
We briefly introduce two popular frameworks in this section:

\begin{itemize}[itemsep=1pt, topsep=1.5pt]
\item \textbf{Bottom-up Hierarchical Vision.} 
    David Marr~\cite{marr2010vision} proposes to break up the vision system into low-level, mid-level, and high-level stages. 
    Low-level vision tasks aim to extract edges, contours, and blobs from raw pixels, while high-level vision concerns the semantics, functionality, and holistic 3D shape of objects. 
    Mid-level vision, lying in-between the low-level and high-level vision, addresses the view-centric geometric properties of surfaces (\eg, depth and surface normal), generic object segmentation and localization \emph{without semantics}, cross-view image correspondences, and so on.
    This framework favors a feed-forward and bottom-up vision process. 
\item 
    \textbf{Three R's of Vision.} Malik~\etal~\cite{malik2016three} categorize vision tasks into recognition, reconstruction, and reorganization. 
    Recognition refers to the task of assigning semantic categories to images.
    Reconstruction concerns the estimation of 3D structures.
    Reorganization addresses the grouping and segmentation of images on spatial or perceptual similarity.
    Instead of a bottom-up process, 
    3R's emphasizes the mutual benefits and relationship across these three visual capabilities.
\end{itemize}

\noindent
Both frameworks highlight the importance of the non-semantic vision tasks --- mid-level vision in Marr's framework, reconstruction and reorganization in 3R's. 
We adopt the term ``mid-level" for the convenience of differentiating from high-level recognition tasks.
However, the underlying vision mechanism remains under debate.

\section{Experimental Setup}
\label{sec:method}

In this section, we first introduce the SSLs considered in this work (Tab.~\ref{tab:ssls}) and our generic evaluation protocols for probing these SSLs. 
We then present the mid-level vision tasks in detail, including their definitions, datasets, and evaluation approaches (Sec.~\ref{subsec:seg}-~\ref{subsec:multiview} and Tab.~\ref{tab:tasks}).

Our probing approaches are largely inspired by Probe3D~\cite{el2024probing}, the recent work that evaluates the 3D awareness of visual foundation models (VFMs).
The VFMs are usually trained under distinct settings (\eg supervision signals, dataset sizes, network architectures and capacity), while we aim at probing self-supervised visual models that are trained with similar configurations.
Therefore, our work provides a complementary study, specifically a more controlled setting, compared to Probe3D~\cite{el2024probing}.
\vspace{-10pt}
\paragraph{Self-Supervised Models.} 
Tab.~\ref{tab:ssls} presents an overview of the SSL models considered in our study.
We primarily focus on SSLs that have publicly available checkpoints pretrained on ImageNet1K~\cite{deng2009imagenet}, with standard neural network architectures (\eg ResNet50~\cite{he2016deep} or Vision Transformers (ViT)~\cite{dosovitskiy2020image}) as the image feature extractor. 
These SSLs cover a wide range of pretraining strategies, including clustering, contrastive learning, generative modeling, or a mixture of different training objectives.
Targeting a comprehensive benchmark, we track the SSL literature back to classic methods that learn image representations from pre-text tasks such as Jigsaw~\cite{Noroozi2016} and RotNet~\cite{Gidaris2018}. 
Our selection of SSLs is inspired by Marks~\etal~\cite{marks2024closer}, which provides a detailed study of the evaluation protocols of SSLs. 

\begin{table}[!tp]
\small
\centering
\begin{tabular}{@{}c p{2.5cm}ccc@{}}
\toprule
\textbf{ID} & \textbf{SSL} & \textbf{Architecture} & \textbf{Category} \\ 
\midrule
1 & Jigsaw~\cite{Noroozi2016}         & ResNet50               & Pre-text          \\
2 & RotNet~\cite{Gidaris2018}         & ResNet50               & Pre-text         \\
3 & NPID~\cite{Wu2018a}              & ResNet50               & Contrastive       \\
4 & NPID++~\cite{misra2019pirl}       & ResNet50               & Contrastive       \\
5 & PIRL~\cite{misra2019pirl}         & ResNet50               & Contrastive     \\
6 & ClusterFit~\cite{yan2019clusterfitimprovinggeneralizationvisual} & ResNet50               & Clustering      \\
7 & SwAV~\cite{caron2021unsupervisedlearningvisualfeatures}        & ResNet50               & Clustering     \\
8 & SeLa-v2~\cite{caron2021unsupervisedlearningvisualfeatures}     & ResNet50               & Clustering      \\
9 & DeepCluster-v2~\cite{caron2021unsupervisedlearningvisualfeatures} & ResNet50            & Clustering       \\
10 & SimCLR~\cite{chen2020simple}      & ResNet50               & Contrastive     \\
11 & MoCo-v2~\cite{chen2020improved}   & ResNet50               & Contrastive     \\
12 & SimSiam~\cite{chen2021exploring}  & ResNet50               & Contrastive      \\
13 & DenseCL~\cite{wang2021dense}      & ResNet50               & Contrastive     \\
14 & BYOL~\cite{grill2020bootstraplatentnewapproach}         & ResNet50  & Contrastive    \\
15 & Barlow Twins~\cite{zbontar2021barlowtwinsselfsupervisedlearning} & ResNet50               &  Contrastive     \\
16 & MoCo-v3~\cite{chen2021empiricalstudytrainingselfsupervised}     & ResNet50                & Contrastive      \\
17 & DINO~\cite{caron2021emerging}         & ResNet50               & Self-distillation      \\
\midrule
18 & MoCo-v3~\cite{chen2021empiricalstudytrainingselfsupervised}     & ViT-B/16                & Contrastive      \\
19 & DINO~\cite{caron2021emerging}         & ViT-B/16               & Self-distillation      \\
20 & MAE~\cite{he2022masked}         & ViT-B/16               & Generative     \\
21 & iBOT~\cite{zhou2021ibot}         & ViT-B/16               & Mixture       \\
22 & MaskFeat~\cite{wei2022masked}    & ViT-B/16               & Generative       \\
\bottomrule
\end{tabular}
\caption{\textbf{Evaluated Self-Supervised Learning Methods (SSLs).} We provide a comprehensive evaluation of SSLs that provide publicly available checkpoints trained from the ImageNet1K dataset~~\cite{deng2009imagenet}.
We categorize SSLs based on their training objectives and network backbones. \textit{Mixture} denotes the combination of different SSLs (\eg, iBOT~~\cite{zhou2021ibot}). 
The SSLs are listed in order of their publication year.
}
\label{tab:ssls}
\end{table}
\vspace{-10pt}
\paragraph{Generic Evaluation Protocols.}
Many mid-level vision tasks can be formulated as dense prediction (\eg, depth estimation).
Unlike image classification, dense prediction requires local pixel or patch representations as input, and  
the main design choice is to decide which network layers (\eg, ResNet layers or ViT blocks) to extract the dense features from.
Following Probe3D, we use the DPT decoder~\cite{ranftl2021vision}, a non-linear dense multi-scale network. 
For ResNet~\cite{he2016deep}, we use the outputs of the last four ResNet blocks as the input to the DPT head, following the architecture of the feature pyramid network~\cite{lin2017feature}; 
For ViT~\cite{dosovitskiy2020image}, following Probe3D, the network is split into four blocks evenly, and the features are extracted after each block --- for ViT-B, we extract features after layers 3, 6, 9, and 12. 
Like Probe3D, we probe the frozen representations to evaluate the capabilities of learned representations instead of their transferability.
We provide task-specific evaluation protocols in the following sections and hyperparameter settings in the supplementary material. 

\begin{table*}[ht!]
\centering
\small
\begin{tabular}{llccccp{2cm}}  
\toprule
\textbf{Hierarchy} & \textbf{Tasks} & \textbf{Type} & \textbf{Benchmarks} & \textbf{Losses} & \textbf{Metrics} \\
\midrule
\multirow{3}{*}{High-level}& Image Classification & 2D & ImageNet1K \cite{imagenet15russakovsky} & CE & Accuracy\\
& Object Detection & 2D & VOC07 \cite{pascal-voc-2007}, COCO \cite{cocodataset} & CE+L2 & AP\\
& Semantic Segmentation & 2D & COCO \cite{cocodataset} & CE & mIOU\\
\arrayrulecolor{gray}\midrule\arrayrulecolor{black}
\multirow{7}{*}{Mid-level} & Generic Object Segmentation & 2D  & VOC07 \cite{pascal-voc-2007}, VOC12 \cite{pascal-voc-2012} & BCE & F1, mIoU, Acc \\
& Depth Prediction& 3D & NAVI \cite{jampani2023navi}, NYU \cite{Silberman:ECCV12} & AdaBin & RMSE, $\epsilon_1, \epsilon_2$, $\epsilon_3$ \\
& Surface Normal Estimation& 3D &  NAVI \cite{jampani2023navi}, NYU \cite{Silberman:ECCV12} & Angular & RMSE, $\epsilon_1$, $\epsilon_2$, $\epsilon_3$ \\
& Geometric Correspondence & 3D & NAVI \cite{jampani2023navi}, ScanNet \cite{dai2017scannetrichlyannotated3dreconstructions} & Training-free & 2D, 3D Recall \\
& Mid-level Image Similarity & 2D  & NIGHTS \cite{fu2024dreamsim} & Training-free & Retrieval Score (F1-Score) \\
\bottomrule
\end{tabular}
\caption{\textbf{Probing Tasks.}
In comparison with the standard SSL evaluation protocols that primarily focus on high-level vision tasks (\textbf{top}), we provide a comprehensive, systematic, and controlled evaluation of SSL in mid-level vision tasks (\textbf{bottom}). Our evaluations cover 2D and 3D vision with a wide range of benchmarks. The $\epsilon_1$, $\epsilon_2$ and $\epsilon_3$ in depth and surface normal estimation represents recall at different threshold rations \cite{eigen2014depthmappredictionsingle, 7410483}. Detailed descriptions of each metric can be found in the supplementary.
}
\label{tab:tasks}
\end{table*}

\subsection{Generic Object Segmentation}
\label{subsec:seg}

Generic object segmentation (or figure-ground segmentation) refers to the task of separating objects (figure) from the surrounding background \emph{without any semantics}. 
This mid-level vision task is different from semantic segmentation (\ie, assigning each pixel to \emph{a semantic category}) which is commonly used for evaluating the high-level visual capabilities of SSLs.

We use the VOC07 and VOC12 datasets~\cite{pascal-voc-2007, pascal-voc-2012} as our benchmarks, following prior research on unsupervised object segmentation~\cite{wang2023cutlearnunsupervisedobject,wang2022freesolo,van2022discovering}.
Given a pretrained feature extractor, we freeze the backbone and train a DPT decoder~\cite{ranftl2021dpt} with a simple binary cross-entropy loss (BCE).
We use standard segmentation metrics (\eg, mean IoU) to measure the performance of SSLs in this task.

\subsection{Monocular Surface Reconstruction}
\label{subsec:3d}

We evaluate SSLs in two tasks related to monocular 2.5D surface reconstruction: depth and surface normal estimation from Probe3D~\cite{el2024probing}.
Even though these two tasks are closely related, they rely on different visual cues and capture distinct geometric properties.

We evaluate depth estimation and surface normal prediction at both scene and object levels. 
For scene-level performance, we employ the NYUv2 dataset \cite{Silberman:ECCV12}, a widely used benchmark for indoor environments. Object-level depth is assessed using the NAVI dataset \cite{jampani2023navi}, which provides diverse object instances across scenes and orientations. 

We present the task-specific configurations as follows.
\vspace{-10pt}
\paragraph{Monocular Depth Estimation} 
The task aims at predicting pixel-wise depth from monocular images. 
We adopt the formulation of binned depth estimation from AdaBins~\cite{Farooq_Bhat_2021}, instead of the classical regression approaches~\cite{Eigen2014DepthMP}. 
We estimate the scale-invariant depth for object-centric images (\eg, NAVI) while predict metric depths on scene-centric datasets (\eg, NYUv2). 
We report root-mean-squared error (RMSE) and recall at various threshold ratios.
\vspace{-10pt}
\paragraph{Surface Normal Estimation.} This task is to predict per-pixel surface direction.
We train the DPT head with the uncertainty-aware angular loss~\cite{bae2021estimatingexploitingaleatoricuncertainty}. 
As for depth estimation, we report RMSE and percentage recall at various angular thresholds.

\subsection{Multiview Correspondence}
\label{subsec:multiview}

In this section, we consider the tasks of estimating correspondences across multiple views that involve mid-level variations, such as changes in camera perspective and object motion. 
These mid-level variations differ from low-level variations in image properties (\eg, color and texture) and high-level variations in semantic attributes (\eg, object categories).
Specifically, we probe SSLs in two tasks: geometric correspondence and mid-level image similarity. Following Probe3D~\cite{el2024probing}, 
we extract image or pixel representations from pretrained SSLs model without additional training.
\vspace{-10pt}
\paragraph{Geometric Correspondence}
This task aims to build pixel correspondences across two views that depict the same 3D scene. 
Our evaluation follows Probe3D~\cite{el2024probing}. Specifically, we estimate the pixel correspondences using the dense feature maps extracted from pretrained SSL models. We use the Paired ScanNet~\cite{dai2017scannet} to evaluate scene-centric images and the NAVI dataset~\cite{jampani2023navi} for object-centric images. 
We sample views with a maximum rotation of 120 degrees to ensure that a mutually visible surface exists without a large viewpoint change. 
We report the correspondence recall --- the percentage of correspondence that falls within some defined distance.

\begin{figure*}[th!]
    \centering
    \includegraphics[width=\linewidth]{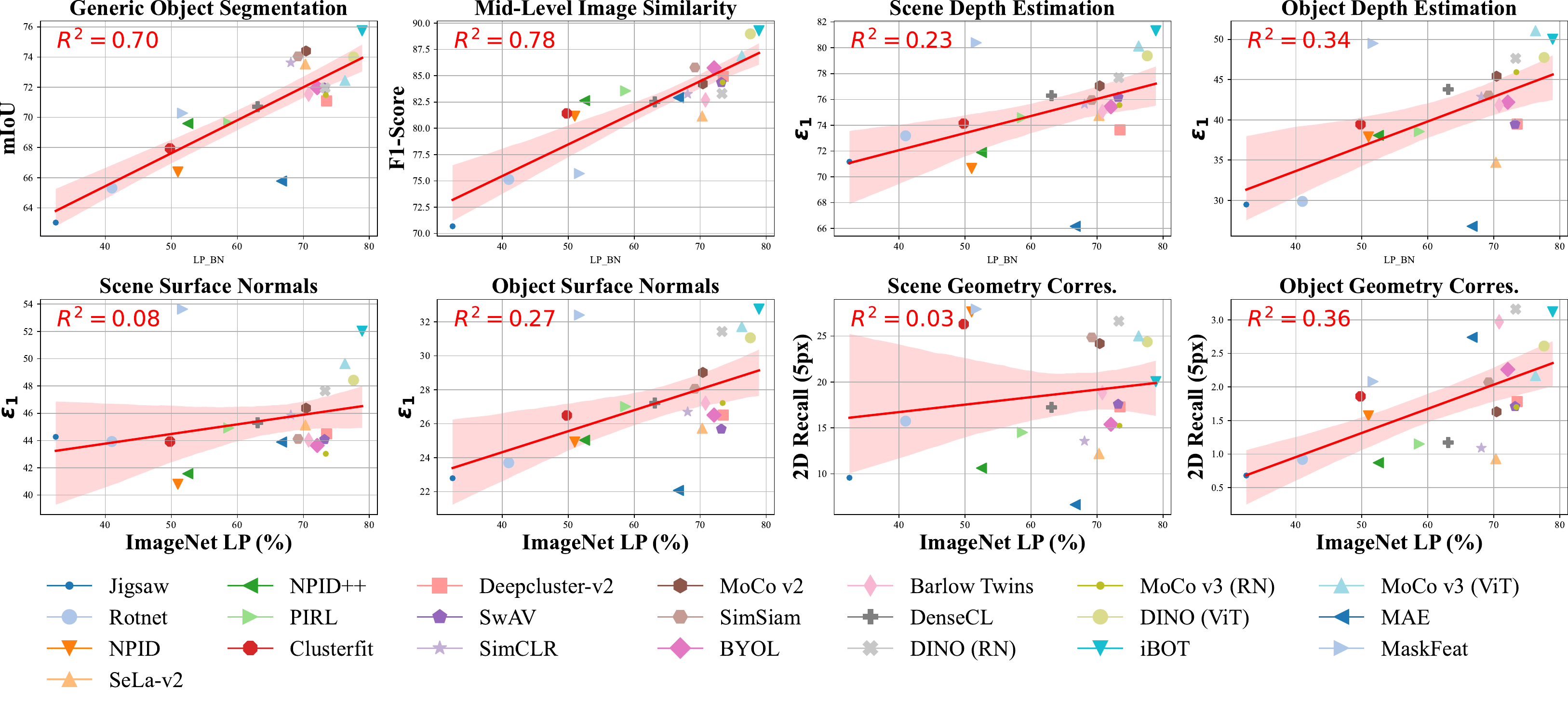}
    \vspace{-6mm}
    \caption{\textbf{Mid-Level Vision \emph{v.s.} ImageNet Linear Probing.} We report the performance of self-supervised learning methods on mid-level vision tasks (y-axis) against their ImageNet 1K linear classification accuracy. Metrics are detailed in Tab.~\ref{tab:tasks}. Linear regression shows correlation with $R^2$ in each plot’s top-left corner, and the red shaded area marks the 95\% confidence interval.
    \vspace{-5px}}    
\label{fig:mid-vs-imagenet}
\end{figure*}
\vspace{-10pt}
\paragraph{Mid-Level Image Similarity}
The task aims at measuring the similarity of two images with mid-level variations (\eg, viewpoint).
We evaluate the SSLs on the NIGHTS dataset collected by DreamSim~\cite{fu2023dreamsimlearningnewdimensions}. 
Specifically, for each triplets $(x, \Tilde{x_0}, \Tilde{x_1})$ from the NIGHTS dataset, we predict whether $\Tilde{x_0}$ or $\Tilde{x_1}$ is more similar to the reference image $x$. 
We denote a distance between two images as $D(x, \tilde{x}; f_\theta) = 1-\text{cos} \big(f_\theta(x), f_\theta(\tilde{x}) \big)$ where $f_{\theta}$ is the image representations from pretrained SSLs.
For ViT, we use the \texttt{CLS} tokens taken from the last layer;
For ResNet, we use the global image representation in the output layer. 
The model’s vote is calculated as follows: if \(d_1 < d_0\), then \(\hat{y} = 1\); otherwise, if \(d_0 < d_1\), then \(\hat{y} = 0\), where \(d_0 = D(x, \tilde{x}_0; f_\theta)\) and \(d_1 = D(x, \tilde{x}_1; f_\theta)\). We use voting accuracy as our evaluation metric.
\section{Analysis}
\label{sec:exp}

\subsection{Does strong high-level performance imply strong mid-level performance?}

\paragraph{Yes, to some extent.} As ImageNet (IN1k) probing results improve, we generally observe a corresponding enhancement in mid-level vision tasks, as depicted in Fig.~\ref{fig:mid-vs-imagenet}. 
This positive trend indicates that SSL models capable of high-level visual recognition tend to provide better representations for mid-level vision tasks as well, benefiting tasks that leverage spatial and structural information.

\paragraph{Some mid-level tasks are highly correlated with high-level.} A per-task closer examination, however, reveals nuances in this relationship. For example, \emph{generic object segmentation} shows the highest correlation with high-level tasks, achieving a coefficient of determination (\(R^2 = 0.70\)). This suggests that high-level features indeed capture spatial structures that enhance segmentation accuracy. 
Mid-level and high-level vision also correlate highly in the task of \emph{mid-level image similarity}. This indicates that the high-level image representations are invariant to mid-level variations to some extent (\eg, viewpoint variations).

\paragraph{3D geometric understanding is weakly correlated with high-level vision.} For tasks like \emph{scene surface normal estimating}, we observe notably weaker correlations. These tasks often require intricate, fine-grained geometric and surface features that high-level SSL objectives alone do not sufficiently capture. This gap implies that mid-level vision depends on specialized cues not fully addressed by high-level feature extraction.

\begin{figure}[t]
    \centering
    \includegraphics[width=\linewidth]{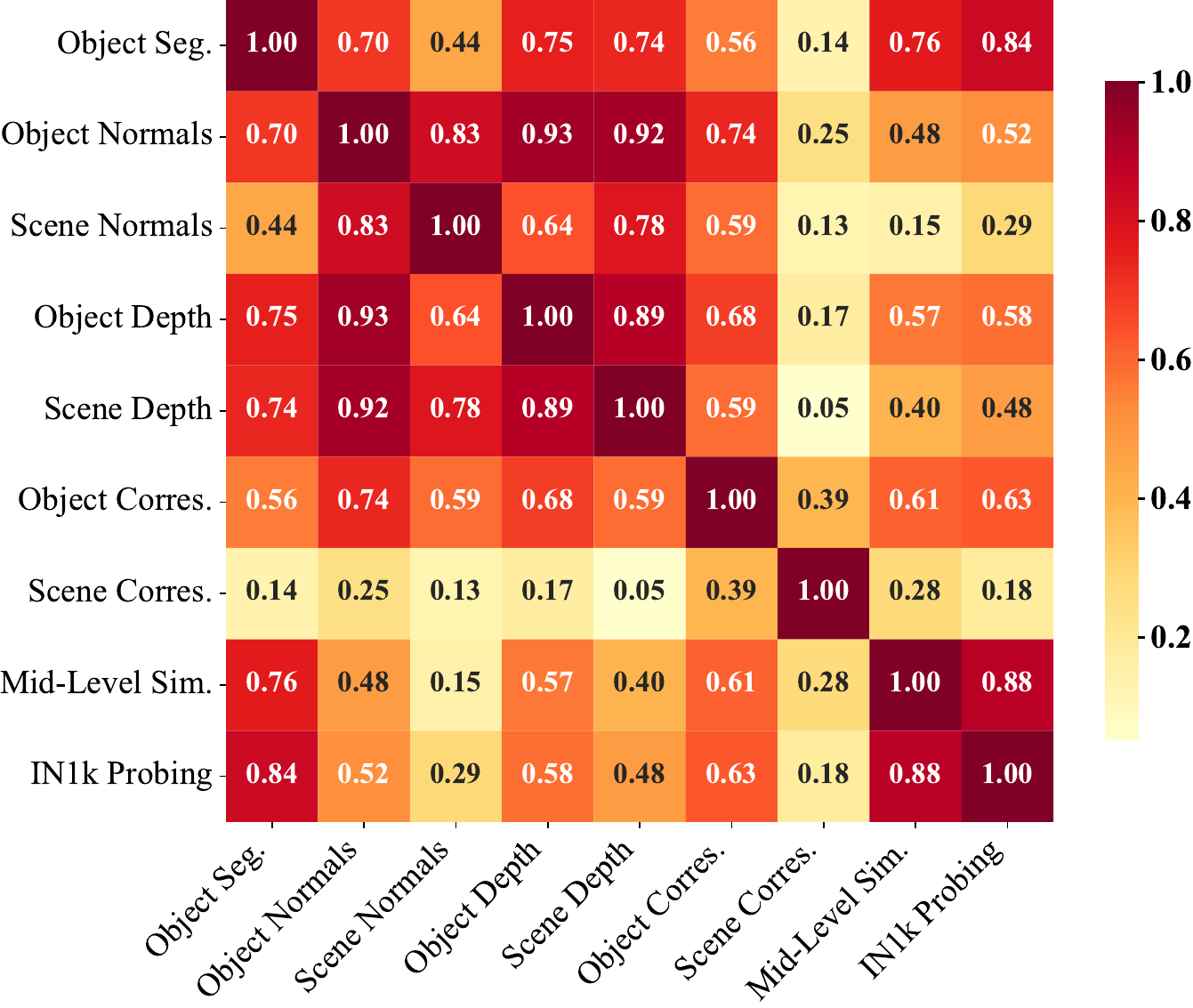}
    \vspace{-8mm}
    \caption{\textbf{Cross-task performance correlation.} We present Pearson coefficients across mid-level and high-level vision tasks. We use the same set of metrics we used in Fig.~\ref{fig:mid-vs-imagenet}.}
    \label{fig:correlation_matrix}
\end{figure}

\paragraph{Correlation across mid-level vision tasks.}
Fig.~\ref{fig:correlation_matrix} shows the Pearson correlations across tasks. 
We find a strong correlation between object segmentation and mid-level image similarity (coefficients above $0.76$).
Object segmentation further correlates well with depth estimation and object-level multi-view geometric correspondence,
indicating that spatial understanding aids segmentation and underscores the importance of 3D awareness in image understanding. 
In contrast, scene geometry correspondence correlates weaker with other mid-level and high-level tasks.
Mid-level image similarity aligns with object segmentation and correspondence but correlates less with 3D tasks. 

\begin{figure}[t!]
\centering
\includegraphics[width=1\linewidth]{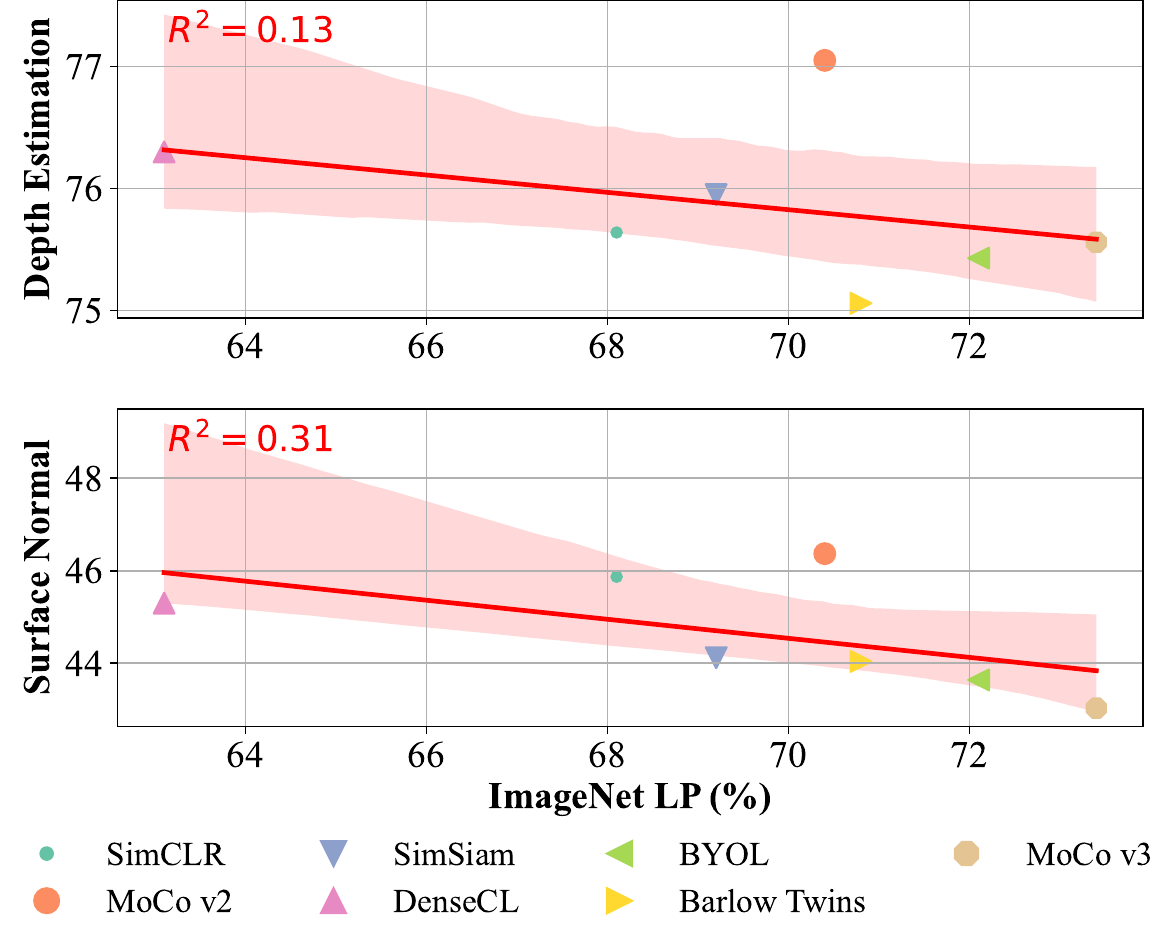}
\caption{\textbf{Mid-Level Vision \emph{v.s.} ImageNet Linear Probing with Contrastive Learning Methods.} We examine several contrastive learning methods~\cite{chen2020simple, chen2021exploring, grill2020bootstraplatentnewapproach, chen2021empiricalstudytrainingselfsupervised, chen2020improved, wang2021dense, zbontar2021barlowtwinsselfsupervisedlearning} and observe a negative correlation between ImageNet linear probing accuracy and task performance. Depth estimation results are shown at the top, with surface normal estimation results at the bottom. The y-axis represents prediction accuracy, where higher values indicate better performance.}
\label{fig:contrastive-learning-impact}
\end{figure}
\vspace{-10pt}
\begin{figure*}[t]
    \centering
    \includegraphics[width=\linewidth]{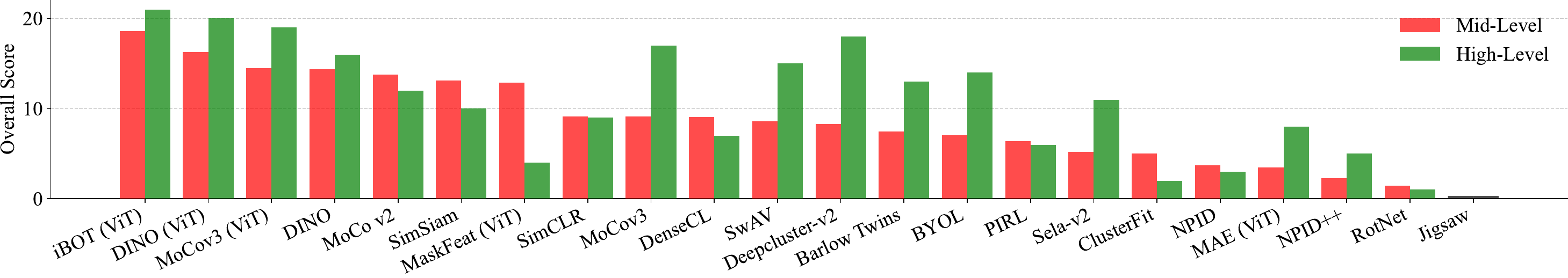}
    \vspace{-6mm}
    \caption{\textbf{Ranking SSLs based on Overall Score.} Comparative ranking of SSL models based on performance in mid-level vision tasks and ImageNet probing. Models are ranked per task, and average scores reflect overall performance across mid-level and high-level vision tasks, with higher scores indicating stronger results.\vspace{-15px}}
    \label{fig:ranking}
\end{figure*}
\paragraph{Ranking in mid-level vision.} In Fig.~\ref{fig:ranking}, we present a comparative ranking of SSL models based on their performance across mid-level vision tasks and ImageNet probing. Each model received a task-specific rank, and we computed an average score, with higher scores indicating stronger performance in both mid-level and high-level vision tasks. Our analysis suggests an imbalance in model performance, with some models excelling in high-level tasks but showing comparatively modest capabilities in mid-level vision tasks.
Notably, MaskFeat stands out as an exception, demonstrating a significant advantage in mid-level vision tasks over high-level performance.

Conversely, MAE ranks lower on mid-level tasks, demonstrating limited proficiency in capturing finer visual details. In the following sections, we will discuss the specifics of MAE and MaskFeat.

Notably, iBOT~\cite{zhou2021ibot} and DINO~\cite{caron2021unsupervisedlearningvisualfeatures} rank among the top performers, exhibiting strong capabilities in both mid-level and high-level vision tasks. In contrast, pretext pre-trained models like RotNet~\cite{Gidaris2018} and Jigsaw~\cite{Noroozi2016} fall behind other SSL methods.

\subsection{What makes a strong mid-level vision model?}
\paragraph{Training objective.}  We observe that the top-performing SSL models in mid-level vision tasks are typically generative-based, such as iBOT~\cite{zhou2021ibot} and MaskFeat~\cite{wei2022masked}. In contrast, recent contrastive self-supervised learning (SSL) models~\cite{chen2020simple, chen2021exploring, grill2020bootstraplatentnewapproach, chen2021empiricalstudytrainingselfsupervised, chen2020improved, wang2021dense, zbontar2021barlowtwinsselfsupervisedlearning} show an inverse relationship between mid-level vision performance and ImageNet linear probing accuracy, as illustrated in Fig.~\ref{fig:contrastive-learning-impact}.

Although these methods have improved ImageNet linear probing accuracy by over 10\%, they exhibit notable declines in mid-level vision tasks. We illustrate this trend with model performance on depth and surface normal estimation and provide a broader analysis across six additional tasks in the supplementary materials. Notably, SSL models pre-trained on pretext tasks show the lowest performance on mid-level vision tasks.
\vspace{-10pt}
\begin{table*}[!ht]
  \centering
  \resizebox{\textwidth}{!}{%
  \setlength\tabcolsep{3pt}
  \small
  \begin{tabular}{l l c c c c c c c c}
    \toprule
    \textbf{SSLs} & \textbf{Backbone} & \textbf{Object Seg.} & \textbf{Object SNorm.} & \textbf{Scene SNorm.} & \textbf{Object Depth} & \textbf{Scene Depth} & \textbf{Object Corres.} & \textbf{Scene Corres.} & \textbf{Image Sim.} \\
    \midrule
    MoCo-v3~\cite{chen2021empiricalstudytrainingselfsupervised} & RN-50    & 71.48 & 27.22 & 43.03 & 45.93 & 75.56 & 1.7 & 15.23 & 84.37 \\
    MoCo-v3~\cite{chen2021empiricalstudytrainingselfsupervised} & ViT-B/16 & \textbf{72.45}& \textbf{31.72} & \textbf{49.64} & \textbf{51.07} & \textbf{80.14} & \textbf{3.12} & \textbf{25.03}  & \textbf{86.90} \\
    \midrule
    DINO~\cite{caron2021emerging}    & RN-50    & 71.95 & \textbf{31.43} & 47.64 & 47.63 & 77.68 & \textbf{3.16} & \textbf{26.63}  & 83.31 \\
    DINO~\cite{caron2021emerging}    & ViT-B/16 & \textbf{74.00} &  31.06 & \textbf{48.42} & \textbf{47.75} & \textbf{79.38} & 2.61 & 24.38  & \textbf{89.36} \\
    \bottomrule
    \toprule
    \textbf{SSLs} & \textbf{Model Size} & \textbf{Object Seg.} & \textbf{Object Norm.} & \textbf{Scene Norm.} & \textbf{Object Depth} & \textbf{Scene Depth} & \textbf{Object Corres.} & \textbf{Scene Corres.} & \textbf{Image Sim.} \\
    \midrule
    iBOT & B/16 & \textbf{75.74} & 32.75 & 52.02 & 50.02 & 81.32 & 3.12 & \textbf{20.04} & \textbf{89.27} \\
    iBOT & L/16 & 75.32 & \textbf{35.65} & \textbf{54.53} & \textbf{52.56} & \textbf{85.14} & \textbf{3.16} & 14.79  & 82.37 \\
    MAE & B/16 & 65.77 & 22.67 & 43.89 & 26.78 & 66.17 & 2.74 & 6.64 & 82.91 \\
    MAE & L/16 & \textbf{66.58} & \textbf{26.6}0 & \textbf{45.60} & \textbf{26.81} & \textbf{70.22} & \textbf{2.93} & \textbf{10.05}  & \textbf{84.26} \\
    \bottomrule
  \end{tabular}
  }
    \caption{\textbf{Impact of Network Architecture and Model Size on Mid-Level Vision Performance.} The top table demonstrates that the Vision Transformer (ViT)~\cite{dosovitskiy2020image} generally outperforms ResNet-50 (RN-50)~\cite{he2016deep} across most mid-level vision tasks when using MoCo-v3 or DINO as the self-supervised pretraining method. The bottom table shows that larger model sizes tend to enhance mid-level vision capabilities. Detailed numerical comparisons are provided in the supplementary material. \vspace{-10px}}
  \label{tab:architecture}
\end{table*}
\paragraph{Network architecture.} 

We further compare ViT-B (ViT Base) and ViT-L (ViT Large) backbones with the same SSL method in Tab.~\ref{tab:architecture}. Our findings show that ViT-L models tend to perform better than ViT-B across mid-level vision tasks. In particular, we evaluate two popular SSL models, iBOT and MAE, on both ViT-B and ViT-L backbones. This pattern suggests that larger backbones like ViT-L may capture mid-level vision features better.

\vspace{-10pt}
\paragraph{Network capability.} 
We also compare ViT-B and ViT-L backbones using the same SSL method in Tab.~\ref{tab:architecture}. Our results indicate that ViT-Large models generally achieve higher scores across various mid-level vision tasks. Specifically, we examine two popular SSL models, iBOT and MAE, with ViT-B and ViT-L backbones. This trend suggests that larger models are better at capturing mid-level vision features.

\subsection{Which SSL works the best per task?}
\begin{table}[ht!]
\small
\centering
\resizebox{\columnwidth}{!}{%
\begin{tabular}{lcccc}
\toprule
\textbf{Model} & \textbf{Image Sim.} & \textbf{Object Corres.} & \textbf{Object Depth} & \textbf{Object SNorm.} \\
\midrule
DINO (RN-50)          & 83.31 & \textbf{3.16} & 47.63 & 31.43 \\
MaskFeat~\cite{wei2022masked} & 75.70 & 2.08 & \underline{49.50} & \underline{32.40} \\
MoCo v3               & \underline{84.37} & 1.70 & 45.93 & 27.22 \\
iBOT~\cite{zhou2021ibot} & \textbf{89.27} & \underline{3.12} & \textbf{50.02} & \textbf{32.75} \\
\bottomrule
\end{tabular}
}

\vspace{2mm} 

\resizebox{\columnwidth}{!}{%
\begin{tabular}{lcccc}
\toprule
\textbf{Model} & \textbf{Object Seg.} & \textbf{Scene Corres.} & \textbf{Scene Depth} & \textbf{Scene SNorm.} \\
\midrule
DINO (RN-50)          & \underline{71.95} & \underline{26.63} & 77.68 & 47.64 \\
MaskFeat~\cite{wei2022masked} & 70.28 & \textbf{27.94} & \underline{80.39} & \textbf{53.63} \\
MoCo v3               & 71.48 & 15.23 & 75.56 & 43.03 \\
iBOT~\cite{zhou2021ibot} & \textbf{75.74} & 20.04 & \textbf{81.32} & \underline{52.02} \\
\bottomrule
\end{tabular}%
}
\caption{\textbf{Top SSL Methods in Mid-Level Vision Tasks.} iBOT outperformed all other SSL methods (listed in Tab.~\ref{tab:ssls}) in multiple mid-level vision tasks. Surprisingly, despite the low performance on ImageNet linear classification, MaskFeat achieved leading results in depth and surface normal estimation.
\vspace{-10px}
}
\label{tab:top_models}
\end{table}

In Tab.~\ref{tab:top_models}, we highlight the best-performing models for each task. iBOT~\cite{zhou2021ibot} consistently leads in most mid-level vision tasks while MaskFeat performs even better on scene correspondence and scene surface normal. DINO works the best on object correspondence and achieves quite competitive results compared to other SSLs.

\subsection{Analysis of specific SSL}
\label{sec:analysis_ssls}
\paragraph{MAE.} MAE~\cite{he2021maskedautoencodersscalablevision} stands out as a clear outlier, as illustrated in Fig.~\ref{fig:mid-vs-imagenet} and Fig.~\ref{fig:ranking}. MAE consistently performs poorly on mid-level vision tasks compared to other ViT-based SSLs such as iBOT~\cite{zhou2021ibot} and MaskFeat~\cite{wei2022masked}.
\vspace{-10pt}

\paragraph{MaskFeat.} In Fig.~\ref{fig:mid-vs-imagenet} and Fig.~\ref{fig:ranking}, we observe that MaskFeat~\cite{wei2022masked} demonstrates surprisingly strong mid-level vision capabilities despite showing relatively modest performance in high-level vision tasks. Since the MaskFeat checkpoint we used was trained exclusively on the ImageNet-1K dataset and followed a training process similar to MAE, we attribute its mid-level vision strength to its choice of target features. Unlike models trained on raw pixels, tokens, or high-level features, MaskFeat is optimized on HOG (Histograms of Oriented Gradients) features, which are widely used for keypoint detection in methods such as SIFT~\cite{SIFT}. Importantly, HOG does not rely on any external models, making it a robust feature for capturing mid-level vision attributes. This finding suggests that incorporating mid-level vision-relevant features, like HOG, into SSL training may enhance model performance on mid-level vision tasks.
\vspace{-10pt}

\paragraph{SSLs based on pre-text tasks.} RotNet and Jigsaw are the two lowest-performing SSLs on mid- and high-level vision. Given that they are the early design of SSLs, it re-validate our claim that ImageNet (IN1k) probing results improve, we generally observe a corresponding enhancement in mid-level vision task peformance.

\subsection{How do visual foundation models (VFMs) perform?}

\begin{figure}[t!]
    \centering
    \includegraphics[width=\linewidth]{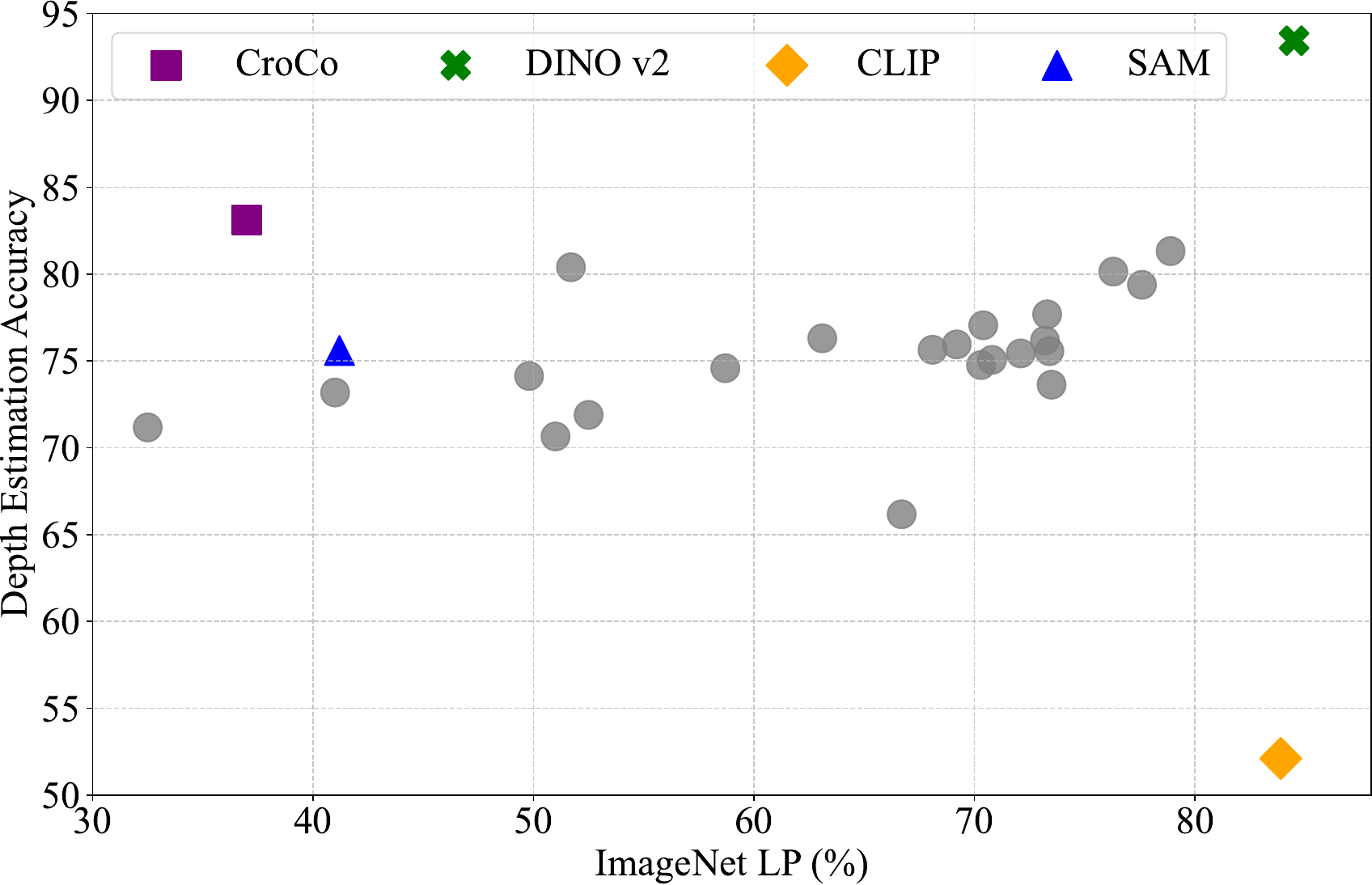}
    \vspace{-6mm}
    \caption{Comparison of ImageNet1K-Pretrained SSLs with models trained with larger datasets (\eg, DINO-v2~\cite{oquab2023dinov2}) or different training supervisions (\eg, CLIP~\cite{radford2021learning}) in the scene depth estimation task. \vspace{-20pt}
}  
\label{fig:vfm}
\end{figure}

Even though our study aims to conduct a controlled benchmarking of SSLs in the mid-level vision tasks, we put VFMs into our context in this section to understand the comparison between the latest visual foundation models and the set of SSLs we analyzed. We observed that CLIP is much worse than most of the SSLs. DINO-v2 is the winner. We include a comparison in Fig.~\ref{fig:vfm} focusing on depth estimation.
\vspace{-10pt}
\paragraph{DINO-V2\textnormal{~\cite{oquab2023dinov2}.}} Our findings, consistent with Probe3D and Cambrian-1~\cite{tong2024cambrian}, show DINO-V2 as the best-performing model across both ImageNet linear probing and mid-level vision tasks, making it a standout choice for comprehensive vision capabilities.
\vspace{-10pt}
\paragraph{CLIP\textnormal{~\cite{radford2021learning}.}} CLIP’s overall performance is notably lower across mid-level tasks, as reported in Probe3D~\cite{el2024probing} and Cambrian-1~\cite{tong2024cambrian}. This suggests that, while generalizable for 2D understanding, CLIP may be less suited for tasks requiring 3D spatial understanding.
\vspace{-10pt}
\paragraph{CroCo\textnormal{~\cite{croco}.}}  Using depth estimation as a representative task, we observe that CroCo performs competitively in mid-level vision tasks despite lagging in high-level vision tasks. Pretrained on synthetic multi-view image datasets, CroCo learns strong 3D-aware image representations. This distinction underscores its potential for 3D-specific applications, where spatial understanding is prioritized over high-level visual semantics because of its design choice.
\vspace{-10pt}
\paragraph{SAM\textnormal{~\cite{kirillov2023segment}}.} Although SAM excels at image segmentation, it exhibits
moderate performance on mid-level vision tasks and notably weaker performance on image classification, in comparison with ImageNet-pretrained SSLs. This reveals the imbalanced performance of SAM among reorganization, reconstruction and recognition.
\section{Discussion}
\label{sec:conclusion}

In conclusion, this study introduces a suite of benchmark protocols to systematically evaluate mid-level vision capabilities, offering a comprehensive and controlled assessment of 22 SSL models across mid-level vision tasks. Our findings indicate a positive correlation between SSL performance in mid-level and high-level vision tasks, though this correlation is generally weaker than that observed within high-level tasks.

Some SSL models exhibit imbalanced performance across these task levels. For instance, MAE~\cite{he2022masked} underperforms in mid-level tasks despite competitive high-level performance, while MaskFeat~\cite{wei2022masked} excels in mid-level tasks relative to its high-level results (Fig.~\ref{fig:ranking}). Mid-level vision capabilities are influenced by factors such as network architecture, dataset size, and pretraining strategy. Overall, ViT~\cite{dosovitskiy2020image} outperforms ResNet~\cite{he2016deep}, increasing model capacity proves beneficial (Tab.~\ref{tab:architecture}), and generative modeling approaches often lead in performance (Fig.~\ref{fig:ranking}). Additionally, we compare SSL models with VFMs, finding that DINOv2~\cite{oquab2023dinov2} leads in both mid-level and high-level vision tasks and language objectives (e.g., CLIP) do not help on mid-level tasks.

Since SSL's overall goal is to produce a very general feature representation that works well for a wide range of downstream tasks, we encourage the community to evaluate new SSL methods on both high- and mid-level tasks.

This work is limited by resources, which prevents us from scaling up our experiments to more tasks and models, 
and we did not include ending tasks such as navigation and robotic manipulation to emphasize mid-level vision’s role further. We provide details of our probing hyper-parameters and include visualizations for each task in the supplementary materials.

\section{Acknowledgement}
\label{sec:ack}

The authors gratefully acknowledge Research Computing at the University of Virginia for providing the computational resources and technical support that made the results in this work possible. \href{https://rc.virginia.edu}{Research Computing at UVA}.

{
    \small
    \bibliographystyle{ieeenat_fullname}
    \bibliography{main}
}

\clearpage
\appendix
\setcounter{page}{1}
\maketitlesupplementary

\begin{table*}[ht]
  \centering
    \caption{
        \textbf{Self-Supervised Model Details.} This table provides details about each model, including the backbone architecture, the dataset used for training, and the source links to the checkpoints utilized in our experiments.
    }
    \label{tab:model-details}
  \setlength\tabcolsep{6pt}
  \footnotesize
  \begin{tabularx}{0.8\linewidth}{|p{5cm}|c|c|X|}
    \toprule
    \textbf{Model Name} & \textbf{Backbone} & \textbf{Dataset} & \textbf{Source Link} \\ 
    \midrule
    Jigsaw~\cite{Noroozi2016} & ResNet-50 & ImageNet-1K & \href{https://github.com/facebookresearch/vissl/blob/main/MODEL_ZOO.md}{VISSL model zoo} \\ 
    RotNet~\cite{Gidaris2018} & ResNet-50 & ImageNet-1K & \href{https://github.com/facebookresearch/vissl/blob/main/MODEL_ZOO.md}{VISSL model zoo} \\ 
    NPID~\cite{Wu2018a} & ResNet-50 & ImageNet-1K & \href{https://github.com/facebookresearch/vissl/blob/main/MODEL_ZOO.md}{VISSL model zoo} \\ 
    SeLa-v2~\cite{caron2021unsupervisedlearningvisualfeatures} & ResNet-50 & ImageNet-1K & \href{https://github.com/facebookresearch/swav}{SwAV repository} \\ 
    NPID++~\cite{misra2019pirl} & ResNet-50 & ImageNet-1K & \href{https://github.com/facebookresearch/vissl/blob/main/MODEL_ZOO.md}{VISSL model zoo} \\ 
    PIRL~\cite{misra2019pirl} & ResNet-50 & ImageNet-1K & \href{https://github.com/facebookresearch/vissl/blob/main/MODEL_ZOO.md}{VISSL model zoo} \\ 
    ClusterFit~\cite{yan2019clusterfitimprovinggeneralizationvisual} & ResNet-50 & ImageNet-1K & \href{https://github.com/facebookresearch/vissl/blob/main/MODEL_ZOO.md}{VISSL model zxwoo} \\ 
    DeepCluster-v2~\cite{caron2021unsupervisedlearningvisualfeatures} & ResNet-50 & ImageNet-1K & \href{https://github.com/facebookresearch/swav}{SwAV repository} \\ 
    SwAV~\cite{caron2021unsupervisedlearningvisualfeatures} & ResNet-50 & ImageNet-1K & \href{https://github.com/facebookresearch/swav}{SwAV repository} \\ 
    SimCLR~\cite{chen2020simple} & ResNet-50 & ImageNet-1K & \href{https://github.com/facebookresearch/vissl/blob/main/MODEL_ZOO.md}{VISSL model zoo} \\ 
    MoCo v2~\cite{chen2020improved} & ResNet-50 & ImageNet-1K & \href{https://github.com/facebookresearch/moco}{MoCo v2 repository} \\ 
    SimSiam~\cite{chen2021exploring} & ResNet-50 & ImageNet-1K & \href{https://mmselfsup.readthedocs.io/en/dev-1.x/model_zoo.html}{MMSelfSup model zoo} \\ 
    BYOL~\cite{grill2020bootstraplatentnewapproach} & ResNet-50 & ImageNet-1K & \href{https://github.com/yaox12/BYOL-PyTorch}{Unofficial BYOL repo} \\ 
    Barlow Twins~\cite{zbontar2021barlowtwinsselfsupervisedlearning} & ResNet-50 & ImageNet-1K & \href{https://mmselfsup.readthedocs.io/en/dev-1.x/model_zoo.html}{MMSelfSup model zoo} \\ 
    DenseCL~\cite{wang2021dense} & ResNet-50 & ImageNet-1K & \href{https://github.com/WXinlong/DenseCL}{DenseCL repository} \\ 
    DINO~\cite{caron2021emerging} & ResNet-50/ViT-B/16 & ImageNet-1K & \href{https://github.com/facebookresearch/dino}{DINO repository} \\ 
    MoCo v3~\cite{chen2021empiricalstudytrainingselfsupervised} & ResNet-50/ViT-B/16 & ImageNet-1K & \href{https://github.com/facebookresearch/moco-v3}{MoCo v3 repository} \\ 
    iBOT~\cite{zhou2021ibot} & ViT-B/16 & ImageNet-1K & \href{https://github.com/bytedance/ibot}{iBOT repository} \\ 
    MAE~\cite{he2021maskedautoencodersscalablevision} & ViT-B/16 & ImageNet-1K & \href{https://github.com/facebookresearch/mae}{MAE repository} \\ 
    MaskFeat~\cite{wei2022masked} & ViT-B/16 & ImageNet-1K & \href{https://mmselfsup.readthedocs.io/en/dev-1.x/model_zoo.html}{MMSelfSup model zoo} \\ 
    \bottomrule
  \end{tabularx}
\end{table*}

Sec.~\ref{sec:ssl-intro} provides an overview the self-supervised learning models (Tab.~\ref{tab:model-details}) included in our study. 
Sec.~\ref{sec:metrics} details the evaluation metrics and presents the quantitative results (Tab.~\ref{tab:app_2d_grouping} -~\ref{tab:image_retrieval_nights}) for each mid-level vision task. 
Sec.~\ref{sec:qualitative} showcases qualitative visualizations (Fig.~\ref{fig:qualitative-depth} -~\ref{fig:qualitative-snorm}). 

\section{Self-supervised Learning Models}
\label{sec:ssl-intro}
In our experiments, we select 22 SSL models from a wide range of categories based on two criteria:
(1) coverage of the main approaches used for large-scale self-supervised training and 
(2) comparable model architecture and training data to allow fair comparisons. 
We primarily evaluate the publicly-available checkpoints pretrained on ImageNet1K~\cite{deng2009imagenet} 
--- the links to each checkpoint are included in Tab.~\ref{tab:model-details}.
We briefly describe each SSL below. 

\paragraph{Jigsaw.}

Noroozi and Favaro \cite{Noroozi2016} introduced a self-supervised learning approach for model pretraining based on solving jigsaw puzzles as a pretext task. This method trains a network to predict the correct arrangement of shuffled image patches, where the image is divided into a 3x3 grid. At its core, this approach encourages the model to learn spatial relationships and understand object structure by generating consistent embeddings for the spatially rearranged patches of the same image. In our study, we used the publicly available ResNet-50 checkpoint trained on the ImageNet-1k~\cite{imagenet15russakovsky}  dataset.

\paragraph{Rotnet.}

Gidaris~\etal~\cite{Gidaris2018} proposed a self-supervised approach for model pretraining using a rotation prediction task, known as RotNet. This method trains a network to classify the rotation angle (0°, 90°, 180°, or 270°) applied to an input image, encouraging the model to learn semantic features and spatial structure within the image. At its core, this approach leverages rotation as a proxy task, pushing the network to recognize objects and their orientations. In our work, we evaluate the ResNet-50 architecture trained on ImageNet-1k~\cite{imagenet15russakovsky} using this pretext task and rely on the checkpoint released by the authors.

\paragraph{NPID.}

Wu~\etal~\cite{Wu2018a} introduced a non-parametric instance-level discrimination approach for unsupervised feature learning. This method trains a network to distinguish between individual instances by treating each image as its own unique class, employing a memory bank to store and update embeddings for all instances in the dataset. At its core, this approach promotes the model to learn discriminative features by maximizing the similarity between augmentations of the same instance and minimizing it across others. In our work, we evaluate the ResNet-50 architecture pre-trained on ImageNet-1k~\cite{imagenet15russakovsky} using this instance discrimination task.

\paragraph{NPID++.} Misra~\etal~\cite{misra2019pirl} significantly improves upon the original implementation of NPID, achieving results that substantially outperform those reported in the original paper~\cite{Wu2018a}.

\paragraph{PIRL.} Misra~\etal~\cite{misra2019pirl} introduced Self-Supervised Learning of Pretext-Invariant Representations (PIRL), a method designed to learn representations that remain invariant across various pretext tasks. The approach applies contrastive learning, where the model is trained to produce similar embeddings for multiple augmentations of the same image while distinguishing between different images. At its core, PIRL combines instance discrimination with pretext invariance to capture both semantic and structural features. In our work, we evaluate the ResNet-50 architecture pre-trained on ImageNet using the PIRL framework.

\paragraph{ClusterFit.} Yan~\etal~\cite{yan2019clusterfitimprovinggeneralizationvisual} proposed ClusterFit, a self-supervised learning approach that improves feature representations through clustering and re-training. This method begins by clustering embeddings of unlabeled images to capture the underlying data distribution, using these cluster assignments as pseudo-labels to retrain the model, thus distilling semantic information at the cluster level. At its core, ClusterFit follows a two-step process—clustering followed by supervised re-training—to develop robust and discriminative features. In our work, we evaluate the checkpoint using ResNet-50 architecture which is pre-trained on ImageNet.

\paragraph{SimCLR.} Chen~\etal~\cite{chen2020simple} proposed SimCLR, a contrastive self-supervised learning framework designed to learn visual representations by maximizing agreement between different augmented views of the same image. The method applies a series of data augmentations, including random cropping, color distortion, and Gaussian blur, and uses a contrastive loss to bring embeddings of the same image instance closer together while pushing apart embeddings of different images. At its core, SimCLR leverages a simple yet effective contrastive objective, removing the need for specialized architectures or memory banks. In our work, we evaluate the ResNet-50 architecture trained on ImageNet-1k~\cite{imagenet15russakovsky}.
 
\paragraph{SwAV.} Caron~\etal~\cite{caron2021unsupervisedlearningvisualfeatures} introduced SwAV (Swapping Assignments between Views), a self-supervised learning approach that combines clustering with contrastive learning. Instead of directly contrasting augmented views, SwAV clusters the features of one view and assigns pseudo-labels, which are then used to predict the cluster assignments of another view. This method enables the model to learn representations without requiring negative samples or a memory bank. At its core, SwAV maximizes similarity between different augmentations by leveraging these swapped cluster assignments. In our work, we evaluate the ResNet-50 architecture trained on ImageNet 1k with SwAV.

\paragraph{SeLa-v2.} SeLa~\cite{asano2020selflabellingsimultaneousclusteringrepresentation} proposes an alternative approach to clustering-based self-supervised learning by formulating the clustering process as an optimization problem. It uses the Sinkhorn-Knopp algorithm to solve this optimization efficiently, ensuring that cluster assignments are balanced across the dataset. This avoids degenerate solutions where all data points are assigned to a single cluster. Caron~\etal~\cite{caron2021unsupervisedlearningvisualfeatures} re-implemented SeLa which improves upon the original SeLa by incorporating additional training improvements introduced in the self-supervised learning literature, such as stronger data augmentation, an MLP projection head, and temperature scaling for contrastive learning and yields better performance.

\paragraph{MoCo-v2.} Chen~\etal~\cite{chen2020improved} proposed MoCo-v2, an improved version of the Momentum Contrast (MoCo) framework for self-supervised learning. MoCo-v2 enhances the original MoCo by incorporating stronger data augmentations (such as color distortion and Gaussian blur) and using an MLP projection head to further improve representation quality. Similar to its predecessor, MoCo-v2 employs a memory bank to maintain a large pool of negative samples and uses a momentum-updated encoder to produce stable representations. At its core, this approach refines instance discrimination with updated augmentations and architecture adjustments. In our work, we evaluate the ResNet-50 architecture trained on ImageNet using MoCo-v2.

\paragraph{SimSiam.} Chen and He \cite{chen2021exploring} proposed SimSiam, a self-supervised learning framework designed to simplify contrastive learning by removing the need for negative samples, momentum encoders, or memory banks. Instead, SimSiam trains a Siamese network with two branches, where one branch predicts the representation of the other. By using only a stop-gradient operation on one branch, SimSiam prevents the network from collapsing to trivial solutions, allowing it to learn meaningful representations from positive pairs alone. At its core, SimSiam is a simple and efficient method that demonstrates the feasibility of contrastive learning without negatives. In our work, we evaluate the ResNet-50 architecture trained on ImageNet 1k with SimSiam.

\paragraph{DenseCL.} Wang~\etal~\cite{wang2021dense} introduced DenseCL, a self-supervised learning approach that extends contrastive learning to dense feature correspondences within images. Unlike traditional contrastive methods focused on global representations, DenseCL aims to learn pixel-level features by contrasting dense local regions between augmented views of the same image. This pixel-level contrastive objective encourages the model to learn spatially detailed representations, which benefit dense prediction tasks such as object detection and segmentation. At its core, DenseCL leverages fine-grained contrastive learning to produce more spatially aware features. In our work, we evaluate the ResNet-50 architecture trained on ImageNet 1k using DenseCL.

\paragraph{BYOL.} Grill~\etal~\cite{grill2020bootstraplatentnewapproach} proposed BYOL, a self-supervised learning framework that learns visual representations without requiring negative samples. BYOL employs two neural networks: a “student” network and a “target” network. The student learns to predict the target’s representation of an augmented view of the same image, and the target network is updated as an exponential moving average of the student. This setup enables the model to avoid trivial solutions by progressively refining representations through self-distillation. At its core, BYOL relies on bootstrap mechanisms and a momentum update to learn meaningful features without contrastive pairs. In our work, we evaluate the ResNet-50 architecture trained on ImageNet 1k using BYOL.

\paragraph{DeepCluster-v2.} Caron~\etal~\cite{caron2019deepclusteringunsupervisedlearning} introduced DeepCluster which uses k-means clustering on deep features to assign pseudo-labels to unlabeled data. These pseudo-labels are then used for training the network in an iterative process. However, DeepCluster suffers from the instability of cluster assignments between epochs, which requires reinitializing the classification layer repeatedly, disrupting the training of the convolutional network.
 Caron~\etal~\cite{caron2021unsupervisedlearningvisualfeatures} re-implement DeepCluster and address ealier issues by introducing explicit comparisons between features and cluster centroids instead of learning a classification layer for cluster assignments. This direct comparison increases the stability and performance of the training process. Additionally, DeepCluster-v2 incorporates modern self-supervised learning tricks and further enhances the method’s performances.

\paragraph{Barlow Twins.} Zbontar~\etal~\cite{zbontar2021barlowtwinsselfsupervisedlearning} proposed Barlow Twins, a self-supervised learning approach designed to reduce redundancy in representations by decorrelating feature dimensions. The method uses a loss function that encourages the cross-correlation matrix between two identical networks’ embeddings of augmented views to be as close to the identity matrix as possible, reducing redundancy across dimensions. This setup allows the model to learn diverse and informative features without the need for negative samples or memory banks. At its core, Barlow Twins promotes redundancy reduction, enhancing feature decorrelation. In our work, we evaluate the ResNet-50 architecture pre-trained on ImageNet 1k using Barlow Twins.

\paragraph{MoCo-v3.} Chen~\etal~\cite{chen2021empiricalstudytrainingselfsupervised} proposed MoCo-v3, an extension of the Momentum Contrast framework tailored for Vision Transformers (ViTs) in self-supervised learning. MoCo-v3 adapts the momentum contrastive learning strategy to ViTs, introducing optimizations such as an MLP projection head and advanced data augmentations. Similar to previous versions, MoCo-v3 leverages a momentum-updated encoder to generate stable features and uses a queue-based memory bank to manage negative samples. At its core, this approach refines contrastive learning by combining MoCo’s momentum mechanism with the ViT architecture. In our work, we evaluate the ViT-B/16 architecture trained on ImageNet using MoCo-v3 and employ the checkpoint released by the authors.

\paragraph{DINO.} Caron~\etal~\cite{caron2021emerging} proposed a self-distillation approach for model pretraining. The proposed approach trains a student network to generate features similar to a teacher network, where the teacher is an exponential moving average
of the student network. At its core, this approach relies on instance discrimination as the model is trained to learn to
generate similar embeddings for different crops of the same image instance. In our work, we evaluate the ViT-B/16 architecture trained on ImageNet-1k. We use the checkpoint released by the authors.

\paragraph{MAE.} He~\etal~\cite{he2021maskedautoencodersscalablevision} showed that training vision transformers to reconstruct images based on randomly masked inputs is an effective pretraining task. Such models are trained with a large masking ratio; e.g., 75\% of the input image patches are masked. In our experiments, we use the ViTB/16 and ViT-L/16 models trained on ImageNet-1k.

\paragraph{MaskFeat.} Wei~\etal~\cite{wei2022masked} introduced MaskFeat, a self-supervised learning approach that learns visual representations by predicting masked visual tokens in videos. MaskFeat leverages a Vision Transformer (ViT) and operates by masking random patches in input video frames, then training the model to predict feature embeddings of these masked regions. This strategy encourages the model to capture rich semantic and spatial features, which generalize well across various downstream tasks. At its core, MaskFeat combines masked prediction with a ViT backbone, making it particularly effective for dense prediction tasks. In our work, we evaluate the ViT-B/16 architecture trained on ImageNet-1k using MaskFeat.

\paragraph{BEiT-v2.} Peng~\etal~\cite{peng2022beitv2maskedimage} proposed BEiT-v2, a self-supervised learning method that improves upon the original BEiT by introducing a more refined tokenization process for masked image modeling. BEiT-v2 leverages a teacher-student framework, where the teacher network generates discrete tokens from image patches, and the student network learns to predict these tokens from masked image patches. This approach enhances the model’s ability to capture fine-grained visual patterns and contextual relationships. At its core, BEiT-v2 combines masked image modeling with a new tokenization strategy to achieve state-of-the-art performance on image classification and downstream tasks. In our work, we evaluate the ViT-B/16 architecture trained on ImageNet-1k using BEiT-v2.

\paragraph{iBOT.} Zhou~\etal~\cite{zhou2021ibot} combine ideas from DINO and MAE by training a model to reconstruct masked dense features based on a teacher network. iBOT uses both an imagelevel and a dense distillation objective. We analyze the ViT-B/16 and ViT-L/16 architectures trained on ImageNet1k and ImageNet-22k. We evaluate the checkpoints released by the authors.

\section{Task-Specific Metric Descriptions}
\label{sec:metrics}

\paragraph{Generic Object Segmentation}

We report the full results in Tab.~\ref{tab:app_2d_grouping} using the following metrics to evaluate generic object segmentation, which involves binary segmentation of foreground objects and background:

\begin{itemize}
    \item \textbf{F1 Score:} The F1 score provides a harmonic mean of precision and recall, offering a balanced evaluation of segmentation performance, particularly in the presence of class imbalance. It is defined as:
    \[
    F1 = \frac{2 \cdot \text{Precision} \cdot \text{Recall}}{\text{Precision} + \text{Recall}}
    \]
    where \textbf{Precision} measures the proportion of correctly predicted foreground pixels among all pixels predicted as foreground, and \textbf{Recall} measures the proportion of correctly predicted foreground pixels relative to all ground truth foreground pixels.

    \item \textbf{Accuracy:} Accuracy quantifies the proportion of correctly classified pixels, encompassing both foreground and background classes. It is defined as:
    \[
    \text{Accuracy} = \frac{\text{Correct Predictions $\cdot$ (Fore. + Back.)}}{\text{Total Pixels}}
    \]
    While simple and intuitive, accuracy may be biased toward the majority class (e.g., background), particularly in cases of class imbalance.

    \item \textbf{Mean Intersection over Union (mIoU):} mIoU assesses segmentation performance by averaging the Intersection over Union (IoU) across all classes (foreground and background). For a given class \( c \), IoU is defined as:
    \[
    \text{IoU}_c = \frac{\text{TP}_c}{\text{TP}_c + \text{FP}_c + \text{FN}_c}
    \]
    where \(\text{TP}_c\), \(\text{FP}_c\), and \(\text{FN}_c\) denote the true positives, false positives, and false negatives for class \( c \). mIoU is computed as:
    \[
    \text{mIoU} = \frac{1}{C} \sum_{c=1}^{C} \text{IoU}_c
    \]
    where \( C = 2 \) for generic object segmentation. mIoU provides a robust evaluation of the model's capacity to capture spatial overlap and resolve fine-grained boundaries.
\end{itemize}

These metrics collectively provide a comprehensive evaluation of the model's performance in binary segmentation tasks, highlighting both pixel-level accuracy and the model's ability to distinguish between foreground and background regions.

\paragraph{Depth Prediction }

We present the complete results for depth prediction in Tab.~\ref{tab:app_depth_ssl}. To evaluate performance, we adopt the setup described in \cite{eigen2014depthmappredictionsingle}, which includes computing the root mean square error (RMSE) and evaluating the prediction accuracy under different threshold criteria. The threshold-based accuracy, denoted as $\delta_i$, measures the proportion of pixels for which the ratio between the predicted depth ($d^{pr}$) and the ground-truth depth ($d^{gt}$) lies below $1.25^i$. Formally, this is defined as:
\begin{align}
\delta_i(d^{pr}, d^{gt}) = \frac{1}{N} \sum_{j=1}^N \Big[ \text{max}\left(\frac{d^{pr}_j}{d^{gt}_j}, \frac{d^{gt}_j}{d^{pr}_j}\right) < 1.25^{i} \Big]
\end{align}
where  $N$  is the total number of pixels,  $d^{pr}$  represents the predicted depth, and  $d^{gt}$  is the ground-truth depth.

\paragraph{Surface Normal Estimation}

For each pixel in the image, the error is defined as the angular deviation (in degrees) between the predicted and ground-truth surface normals. To evaluate the model’s performance, we compute two primary metrics: (1) the root mean square error (RMSE), which measures the overall angular error, and (2) the accuracy of predictions at predefined angular thresholds. Specifically, the accuracy metric is calculated as the proportion of pixels whose angular error falls within thresholds of $11.25^\circ$, $22.5^\circ$, and $30^\circ$, following established evaluation protocols~\cite{bae2021estimatingexploitingaleatoricuncertainty,piccinelli2023idisc,7410483}.

\paragraph{Geometric Correspondence}

We report full results on object geometric correspondence in Tab.~\ref{tab:navi_geometric_correspondence} and scene geometric correspondence in Tab.~\ref{tab:scannet_geometric_correspondence}. Correspondences are evaluated using either 2D projection error or 3D metric error. For a correspondence between pixel locations \(p\) in image 1 and \(q\) in image 2, the 2D projection error is computed as follows. First, \(p\) is projected into 3D space, yielding a 3D point \(\mathbf{P}\), using the depth value at \(p\) and the camera intrinsics of image 1. The 3D point \(\mathbf{P}\) is transformed to the coordinate frame of image 2 using the relative camera pose and projected back onto the image plane of image 2, yielding the pixel location \(p'\). The 2D projection error is then defined as:
\[
\text{Error}_{2D} = \| p' - q \|_2
\]
where \(\| \cdot \|_2\) represents the Euclidean distance in the image plane. 

For 3D metric error, both \(p\) and \(q\) are transformed into a shared 3D coordinate space, resulting in \(\mathbf{P}\) and \(\mathbf{Q}\), respectively. The 3D metric error is then computed as:
\[
\text{Error}_{3D} = \| \mathbf{P} - \mathbf{Q} \|_2
\]
The 2D projection error is used for scene-level correspondences, while the 3D metric error is preferred for objects to better account for occlusions and thin structures.

To evaluate correspondence quality, we compute \textit{correspondence recall}, defined as the percentage of correspondences with error below a threshold \(\tau\):
\[
\text{Recall} = \frac{\lvert \{ \text{Error} < \tau \} \rvert}{N}
\]

Where \( \lvert \{ \text{Error} < \tau \} \rvert \) indicates the number of correspondences with error below the threshold \( \tau \) and \( N \) is the total number of correspondences. We report recall values for various \(\tau\) values and analyze results across image pairs grouped by relative viewpoint changes.

\paragraph{Mid-level Image Similarity}

We present the full results for mid-level image similarity in Tab.~\ref{tab:image_retrieval_nights}. In this task, a reference image is provided, and the model selects one of two candidate images based on mid-level image similarity. The evaluation metrics used are Accuracy (Acc), Precision (Prec), Recall (Rec), and F1 Score (F1), defined as follows:

\noindent \textbf{Accuracy (Acc):} The proportion of correctly predicted matches out of the total comparisons:
\[
\text{Acc} = \frac{\text{Correct Predictions}}{\text{Total Comparisons}}
\]

\noindent \textbf{Precision (Prec):} The proportion of correctly identified matches (true positives, TP) among all images predicted as matches:
\[
\text{Prec} = \frac{\text{TP}}{\text{TP} + \text{False Positives (FP)}}
\]

\noindent \textbf{Recall (Rec):} The proportion of correctly identified matches (TP) among all actual matches in the dataset:
\[
\text{Rec} = \frac{\text{TP}}{\text{TP} + \text{False Negatives (FN)}}
\]

\noindent \textbf{F1 Score (F1):} The harmonic mean of Precision and Recall, providing a balanced measure of performance:
\[
\text{F1} = \frac{2 \cdot \text{Prec} \cdot \text{Rec}}{\text{Prec} + \text{Rec}}
\]

These metrics provide a rigorous evaluation of the model's ability to identify mid-level image similarities accurately and consistently.

\section{Qualitative Comparisons}
\label{sec:qualitative}
\begin{figure*}[ht!]
    \centering
    \includegraphics[width=0.9\linewidth]{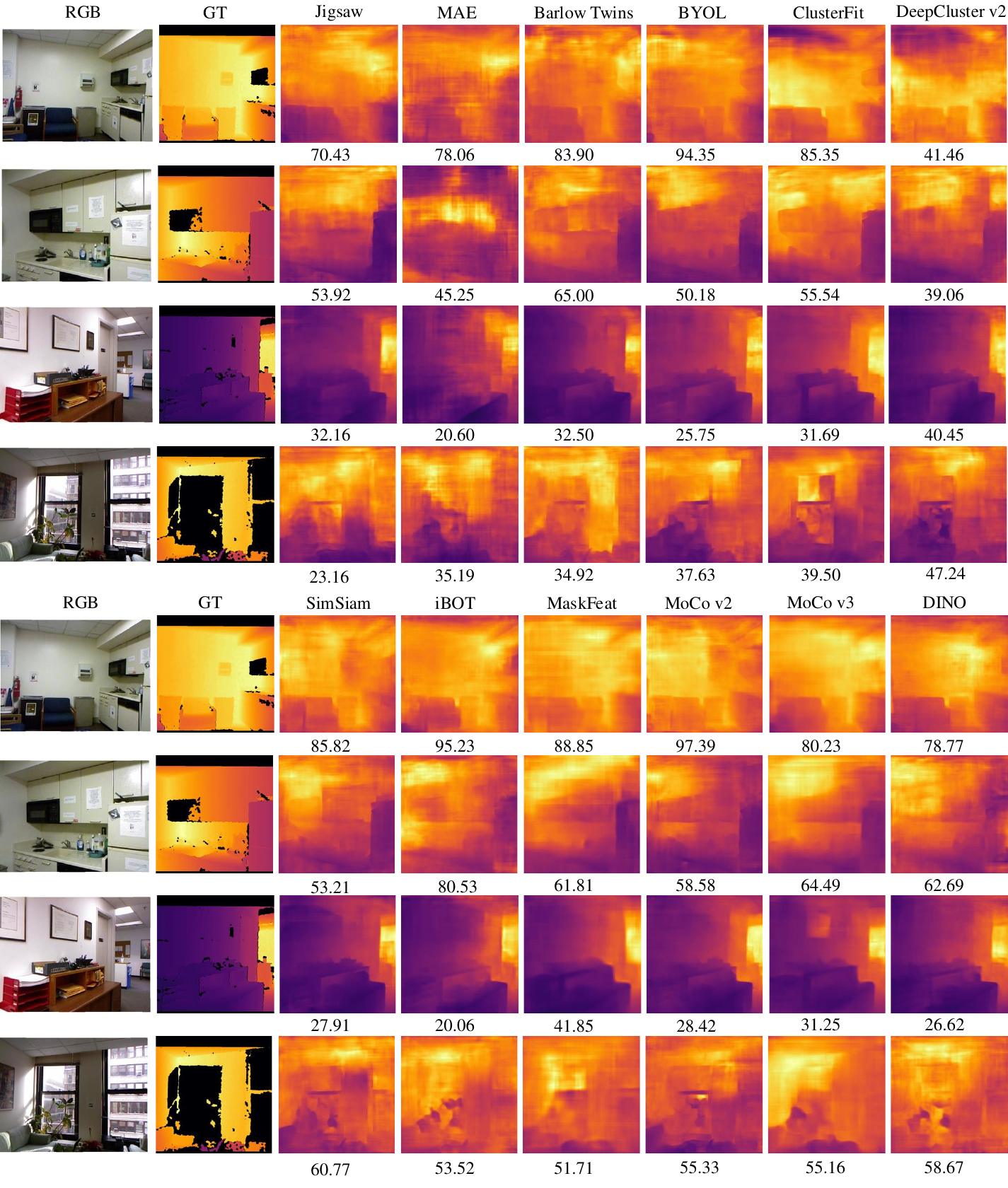}
    \caption{\textbf{Qualitative Depth Estimation Results for Selected SSL Models.} Depth estimation visualizations are shown for selected SSL models, with the $\delta_1$ score displayed below each visualization (\textit{higher is better}). These results highlight the models' effectiveness in capturing depth information. Note DINO and MoCo v3 are ViT based.}
\label{fig:qualitative-depth}
\end{figure*}

\begin{figure*}[ht!]
    \centering
    \includegraphics[width=0.9\linewidth]{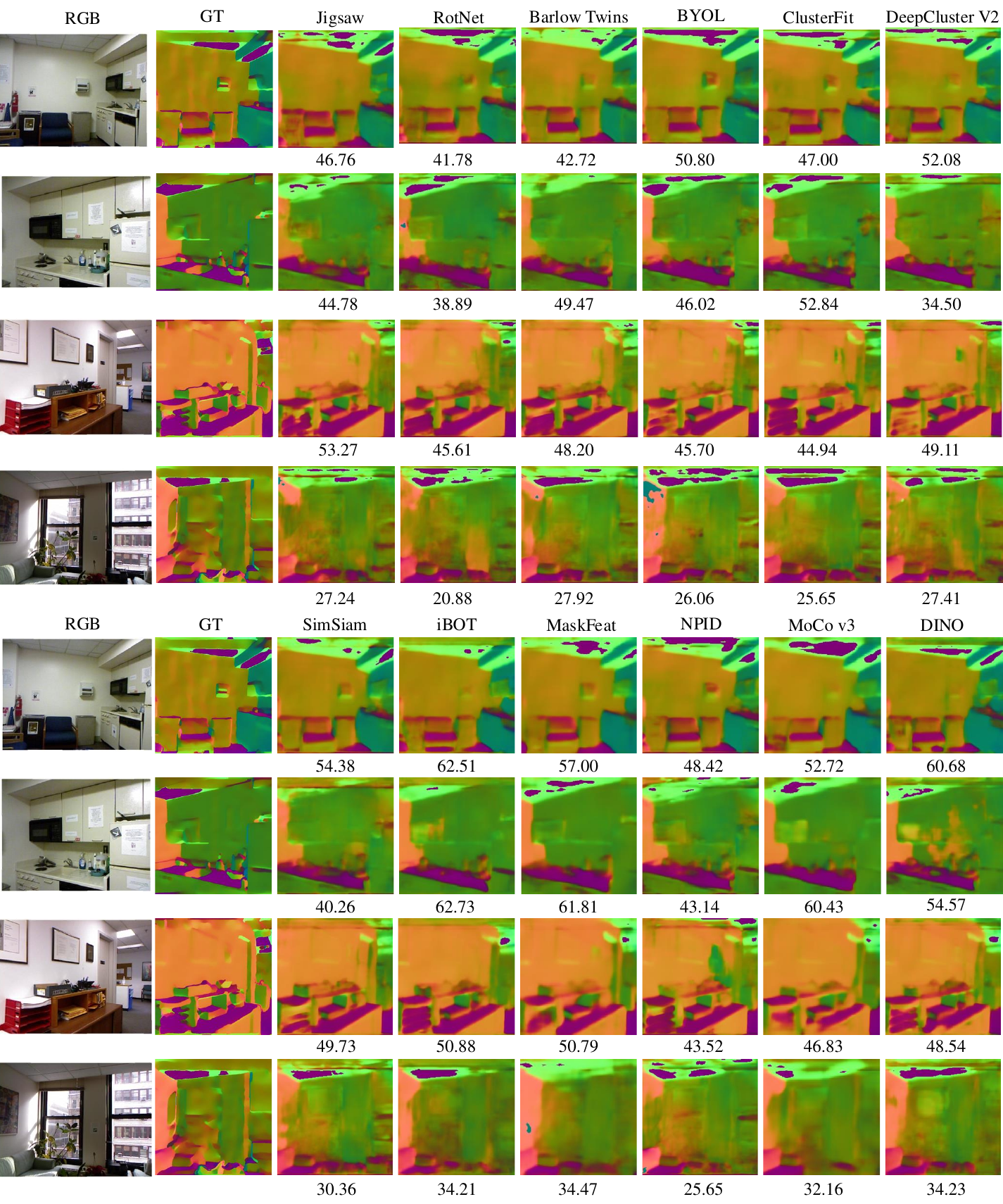}
    \caption{\textbf{Qualitative Surface Normal Estimation Results for Selected SSL Models.} Surface normal estimation visualizations are shown for selected SSL models, with the $\delta_1$ score displayed below each visualization (\textit{higher is better}). These results highlight the models' effectiveness in capturing surface normal information. Note DINO and MoCo v3 are ViT based.}
\label{fig:qualitative-snorm}
\end{figure*}

We present qualitative visualizations in Fig.~\ref{fig:qualitative-depth} and Fig.~\ref{fig:qualitative-snorm} to assess model performance on mid-level vision tasks. These visualizations validate the models' ability to learn and perform each mid level vision task effectively.

\begin{table*}[th!]
  \newcommand{\rowheader}{\rowcolor{Gray!20}}
  \centering
    \caption{
        \textbf{2D Grouping Results (Generic Object Segmentation).} Evaluation results for generic object segmentation, where models segment foreground objects from the background, are presented for both VOC07~\cite{pascal-voc-2007} and VOC12~\cite{pascal-voc-2012} datasets.
    }
    \label{tab:app_2d_grouping}
  \setlength\tabcolsep{6pt}
  \footnotesize
  \begin{tabularx}{\linewidth}{Xll ccc ccc}
    \toprule
    & &
    & \multicolumn{3}{c}{VOC07~\cite{pascal-voc-2007}}  
    & \multicolumn{3}{c}{VOC12~\cite{pascal-voc-2012}} 
    \\
    \cmidrule(lr){4-6} 
    \cmidrule(lr){7-9}
    \textbf{Model} & \textbf{Backbone} & \textbf{Task} &
    F1-measure & mIoU & Accuracy &
    F1-measure & mIoU & Accuracy \\
    \midrule
\midrule \rowheader \multicolumn{9}{l}{\textit{\textbf{Self-Supervised Models (SSL)}}} \\
Jigsaw~\cite{Noroozi2016}          & RN-50      & IN-1k & 71.13 & 63.03 & 83.24 & 81.51 & 71.48 & 89.41 \\
RotNet~\cite{Gidaris2018}          & RN-50      & IN-1k & 75.84 & 65.32 & 85.39 & 83.46 & 71.46 & 89.94 \\
NPID~\cite{Wu2018a}                & RN-50      & IN-1k & 76.92 & 66.38 & 85.99 & 84.34 & 72.66 & 90.35 \\
SeLa-v2~\cite{caron2021unsupervisedlearningvisualfeatures}         & RN-50      & IN-1k & 83.20 & 73.53 & 89.73 & 86.03 & 76.56 & 91.71 \\
NPID++~\cite{misra2019pirl}         & RN-50      & IN-1k & 80.75 & 69.59 & 87.84 & 85.46 & 75.24 & 91.29 \\
PIRL~\cite{misra2019pirl}           & RN-50      & IN-1k & 79.55 & 69.62 & 87.69 & 86.40 & 77.39 & 92.46 \\
ClusterFit~\cite{yan2019clusterfitimprovinggeneralizationvisual}      & RN-50      & IN-1k & 77.91 & 67.94 & 86.79 & 85.58 & 72.98 & 90.25 \\
DeepCluster-v2~\cite{caron2021unsupervisedlearningvisualfeatures} & RN-50      & IN-1k & 79.33 & 71.08 & 88.14 & 88.29 & 79.91 & 93.01 \\
SwAV~\cite{caron2021unsupervisedlearningvisualfeatures}           & RN-50      & IN-1k & 79.72 & 71.95 & 88.59 & 87.38 & 78.72 & 92.91 \\
SimCLR~\cite{chen2020simple}          & RN-50      & IN-1k & 81.05 & 73.63 & 89.44 & 87.94 & 79.62 & 93.25 \\
MoCo v2~\cite{chen2020improved}       & RN-50      & IN-1k & 82.78 & 74.40 & 89.91 & 88.65 & 79.75 & 93.21 \\
SimSiam~\cite{chen2021exploring}       & RN-50      & IN-1k & 82.99 & 74.05 & 89.88 & 88.25 & 77.51 & 92.05 \\
BYOL~\cite{grill2020bootstraplatentnewapproach}       & RN-50      & IN-1k & 83.20 & 71.97 & 89.21 & 87.74 & 78.81 & 93.09 \\
Barlow Twins~\cite{zbontar2021barlowtwinsselfsupervisedlearning} & RN-50      & IN-1k & 79.97 & 71.53 & 88.51 & 88.09 & 78.62 & 92.82 \\
DenseCL~\cite{wang2021dense}          & RN-50      & IN-1k & 79.32 & 70.71 & 88.03 & 87.19 & 78.75 & 92.47 \\
DINO~\cite{caron2021emerging}          & RN-50      & IN-1k & 78.13 & 71.95 & 88.32 & 88.81 & 79.86 & 92.99 \\
MoCo v3~\cite{chen2021empiricalstudytrainingselfsupervised}       & RN-50      & IN-1k & 82.56 & 71.48 & 88.88 & 85.44 & 77.41 & 92.06 \\
DINO~\cite{caron2021emerging}          & ViT-B/16   & IN-1k & 83.12 & 74.00 & 89.79 & 88.70 & 79.94 & 93.17 \\
iBOT~\cite{zhou2021ibot}           & ViT-B/16   & IN-1k & 82.85 & 75.74 & 90.50 & 90.51 & 84.72 & 94.90 \\
MoCo v3~\cite{chen2021empiricalstudytrainingselfsupervised}       & ViT-B/16   & IN-1k & 80.92 & 72.45 & 88.99 & 82.11 & 74.11 & 90.71 \\
MAE~\cite{he2021maskedautoencodersscalablevision}         & ViT-B/16   & IN-1k & 77.25 & 65.78 & 85.88 & 80.22 & 69.63 & 89.14 \\
MaskFeat~\cite{wei2022masked}         & ViT-B/16   & IN-1k & 78.84 & 70.28 & 87.76 & 84.27 & 75.14 & 91.00 \\
\bottomrule
\end{tabularx}
\end{table*}

\begin{table*}[th!]
  \newcommand{\rowheader}{\rowcolor{Gray!20}}
  \centering
\caption{
    \textbf{Depth Estimation Results for SSL Models on NYU and NAVI.} Results for scene-level (NYU) and object-level (NAVI) depth estimation using self-supervised models. These results demonstrate the performance of SSL models across diverse depth estimation tasks.
}
    \label{tab:app_depth_ssl}
  \setlength\tabcolsep{6pt}
  \footnotesize
  \begin{tabularx}{\linewidth}{Xll cccc cccc}
    \toprule
    & &
    & \multicolumn{4}{c}{NYU}  
    & \multicolumn{4}{c}{NAVI} 
    \\
    \cmidrule(lr){4-7} 
    \cmidrule(lr){8-11}
    \textbf{Model} & \textbf{Architecture} & \textbf{Dataset} &
    $\delta_1$ & $\delta_2$ & $\delta_3$ & RMSE &
    $\delta_1$ & $\delta_2$ & $\delta_3$ & RMSE \\
    \midrule
\midrule \rowheader \multicolumn{11}{l}{\textit{\textbf{Self-Supervised Models}}} \\
Jigsaw~\cite{Noroozi2016}                 & RN-50   & IN-1k     &  71.17 &  93.02 &  98.24 & 0.6282    &  29.48 &  55.45 &  73.66 & 0.1775 \\
RotNet~\cite{Gidaris2018}                 & RN-50   & IN-1k     &  73.18 &  93.41 &  98.23 & 0.6047    &  29.87 &  55.03 &  73.00 & 0.1804 \\
NPID~\cite{Wu2018a}                     & RN-50   & IN-1k     &  70.65 &  92.81 &  98.34 & 0.6191    &  37.88 &  65.46 &  80.82 & 0.1506 \\
Sela-v2~\cite{caron2021unsupervisedlearningvisualfeatures}                  & RN-50   & IN-1k     &  74.76 &  94.47 &  98.80 & 0.5684    &  34.72 &  61.97 &  78.64 & 0.1586 \\
NPID++~\cite{misra2019pirl}                 & RN-50   & IN-1k     &  71.89 &  93.27 &  98.34 & 0.6110    &  38.07 &  65.32 &  80.69 & 0.1525\\
PIRL~\cite{misra2019pirl}                     & RN-50   & IN-1k     &  74.58 &  94.13 &  98.59 & 0.5780    &  38.55 &  65.36 &  80.86 & 0.1495 \\
ClusterFit~\cite{yan2019clusterfitimprovinggeneralizationvisual}         & RN-50   & IN-1k     &  74.13 &  93.81 &  98.25 & 0.5850    &  39.45 &  66.47 &  81.45 & 0.1479 \\
DeepCluster-v2~\cite{caron2021unsupervisedlearningvisualfeatures}  & RN-50   & IN-1k     &  73.63 &  93.62 &  98.39 & 0.5863    &  39.50 &  67.35 &  82.43 & 0.1448 \\
SwAV~\cite{caron2021unsupervisedlearningvisualfeatures}                     & RN-50   & IN-1k     &  76.17 &  94.96 &  98.81 & 0.5542    &  39.45 &  67.13 &  82.04 & 0.1457 \\
SimCLR~\cite{chen2020simple}                 & RN-50   & IN-1k     &  75.64 &  94.67 &  98.65 & 0.5698    &  42.86 &  70.04 &  83.68 & 0.1365 \\
MoCo v2~\cite{chen2020improved}                & RN-50   & IN-1k     &  77.05 &  94.83 &  98.77 & 0.5467    &  45.42 &  72.55 &  85.42 & 0.1309 \\
SimSiam~\cite{chen2021exploring}               & RN-50   & IN-1k     &  75.95 &  94.74 &  98.78 & 0.5628    &  43.03 &  70.01 &  83.94 & 0.1366 \\
BYOL~\cite{grill2020bootstraplatentnewapproach}                     & RN-50   & IN-1k     &  75.43 &  94.48 &  98.68 & 0.5711    & 42.19 &  69.22 &  83.54 & 0.1387 \\
Barlow Twins~\cite{zbontar2021barlowtwinsselfsupervisedlearning}      & RN-50   & IN-1k     &  75.06 &  94.22 &  98.61 & 0.5791    &  41.83 &  68.74 &  83.01 & 0.1408 \\
DenseCL~\cite{wang2021dense}               & RN-50   & IN-1k     &  76.30 &  94.69 &  98.65 & 0.5615    &  43.78 &  71.45 &  85.01 & 0.1332 \\
DINO~\cite{caron2021emerging}                     & RN-50   & IN-1k     &  77.68 &  95.89 &  99.09 & 0.5235    &  47.63 &  74.31 &  86.54 & 0.1241 \\
MoCo v3~\cite{chen2021empiricalstudytrainingselfsupervised}                & RN-50   & IN-1k     &  75.56 &  94.63 &  98.86 & 0.5584    &  45.93 &  72.87 &  85.57 & 0.1309 \\
DINO~\cite{caron2021emerging}                     & ViT-B/16 & IN-1k    &  79.38 &  95.97 &  99.05 & 0.5278    &  47.75 &  74.65 &  87.02 & 0.1241 \\
iBOT~\cite{zhou2021ibot}                     & ViT-B/16 & IN-1k    &  81.32 &  96.90 &  99.34 & 0.4919    &  50.02 &  76.29 &  87.89 & 0.1199 \\
MoCo v3~\cite{chen2021empiricalstudytrainingselfsupervised}                & ViT-B/16 & IN-1k    &  80.14 &  96.14 &  99.16 & 0.5109    &  51.07 &  76.96 &  87.95 & 0.1175 \\
MAE~\cite{he2021maskedautoencodersscalablevision}                       & ViT-B/16 & IN-1k    &  66.17 &  90.38 &  97.37 & 0.6898    &  26.78 &  51.82 &  71.69 & 0.1868\\
MaskFeat~\cite{wei2022masked}             & ViT-B/16 & IN-1k    &  80.39 &  96.18 &  99.07 & 0.5125    &  49.50 &  75.47 &  87.14 & 0.1195 \\
\bottomrule
\end{tabularx}
\end{table*}

\begin{table*}[!ht]
\newcommand{\rowheader}{\rowcolor{Gray!20}}
  \centering
    \caption{\textbf{Surface Normal Estimation Results on NYUv2 and NAVI Datasets.} Performance of self-supervised models on scene-level (NYUv2) and object-level (NAVI) surface normal estimation, evaluated using angular thresholds (11.25°, 22.5°, 30°) and RMSE metrics.}
  \label{tab:app_snorm}
  \setlength\tabcolsep{6pt}
  \footnotesize
  \begin{tabularx}{\linewidth}{Xll cccc cccc}
    \toprule
    & &
    & \multicolumn{4}{c}{NYUv2}  
    & \multicolumn{4}{c}{NAVI} 
    \\
    \cmidrule(lr){4-7} 
    \cmidrule(lr){8-11}    
    \textbf{Model} & \textbf{Backbone} & \textbf{Dataset} &
    \textbf{11.25°} & \textbf{22.5°} & \textbf{30°} & \textbf{RMSE} &
    \textbf{11.25°} & \textbf{22.5°} & \textbf{30°} & \textbf{RMSE} \\
    \midrule \rowheader \multicolumn{11}{l}{\textit{\textbf{Self-Supervised Models}}} \\
    Jigsaw~\cite{Noroozi2016}          & RN-50    & IN-1k   & 44.27\ & 67.65\ & 76.23\ & 28.8386 & 22.79\ & 49.22\ & 62.50\ & 36.6169 \\
    RotNet~\cite{Gidaris2018}          & RN-50    & IN-1k   & 43.93\ & 67.40 \ & 76.07\ & 28.8557 & 23.70\ & 50.20\ & 63.46\ & 28.8557 \\
    NPID~\cite{Wu2018a}            & RN-50    & IN-1k   & 40.80\ & 64.68\ & 73.97\ & 35.4511 & 24.92\ & 51.82\ & 64.87\ & 35.4511 \\ 
    SeLa-v2~\cite{caron2021unsupervisedlearningvisualfeatures}         & RN-50    & IN-1k   & 45.14\ & 68.98\ & 77.53\ & 28.0449 & 25.73\ & 53.19\ & 66.22\ & 34.7204 \\
    NPID++~\cite{misra2019pirl}           & RN-50    & IN-1k   & 41.57\ & 65.98\ & 75.14\ & 29.2829 & 25.03\ & 52.03\ & 65.20\ & 34.9940 \\
    PIRL~\cite{misra2019pirl}             & RN-50    & IN-1k   & 44.92\ & 68.35\ & 76.71\ & 28.5771 & 27.01\ & 54.06\ & 66.85\ & 34.1514 \\
    ClusterFit~\cite{yan2019clusterfitimprovinggeneralizationvisual}      & RN-50    & IN-1k   & 43.93\ & 67.40\ & 76.12\ & 28.9261 & 25.49\ & 53.21\ & 65.98\ & 34.8134 \\
    Deepcluster-v2~\cite{caron2021unsupervisedlearningvisualfeatures}  & RN-50    & IN-1k   & 44.48\ & 68.29\ & 76.98\ & 28.2509 & 26.51\ & 54.01\ & 67.07\ & 34.1514 \\
    SwAV~\cite{caron2021unsupervisedlearningvisualfeatures}            & RN-50    & IN-1k   & 44.08\ & 67.98\ & 76.81\ & 28.2881 & 25.69\ & 53.17\ & 66.21\ & 34.4863 \\
    SimCLR~\cite{chen2020simple}          & RN-50    & IN-1k   & 45.87\ & 69.17\ & 77.48\ & 27.9438 & 26.70\ & 54.21\ & 67.07\ & 34.1743 \\  
    MoCo v2~\cite{chen2020improved}         & RN-50    & IN-1k   & 46.37\ & 69.79\ & 78.03\ & 27.5874 & 29.02\ & 56.86\ & 69.42\ & 32.7033 \\
    SimSiam~\cite{chen2021exploring}         & RN-50    & IN-1k   & 44.12\ & 67.95\ & 76.72\ & 28.4032 & 28.06\ & 55.71\ & 68.18\ & 33.5474 \\
    BYOL~\cite{grill2020bootstraplatentnewapproach}            & RN-50    & IN-1k   & 43.64\ & 67.73\ & 76.46\ & 28.5432 & 26.51\ & 54.29\ & 67.17\ & 34.1015 \\
    Barlow Twins~\cite{zbontar2021barlowtwinsselfsupervisedlearning}    & RN-50    & IN-1k   & 44.04\ & 67.75\ & 76.57\ & 28.4161 & 27.21\ & 54.70\ & 67.46\ & 33.9390 \\ 
    DenseCL~\cite{wang2021dense}         & RN-50    & IN-1k   & 45.30\ & 68.74\ & 77.16\ & 28.2974 & 27.21\ & 54.70\ & 67.46\ & 33.9390 \\ 
    DINO~\cite{caron2021emerging}            & RN-50    & IN-1k   & 47.64\ & 70.96\ & 79.12\ & 26.8891 & 31.43\ & 59.50\ & 71.77\ & 31.3895 \\
    MoCo v3~\cite{chen2021empiricalstudytrainingselfsupervised}         & RN-50    & IN-1k   & 43.03\ & 67.15\ & 76.20\ & 28.6994 & 27.22\ & 55.03\ & 67.85\ & 33.7240 \\
    DINO~\cite{caron2021emerging}            & ViT-B/16 & IN-1k   & 48.42\ & 69.71\ & 77.57\ & 28.0873 & 31.66\ & 58.58\ & 70.68\ & 31.9912 \\
    iBOT~\cite{zhou2021ibot}            & ViT-B/16 & IN-1k   & 52.02\ & 72.43\ & 79.53\ & 26.9539 & 32.75\ & 60.06\ & 71.69\ & 31.4563 \\ 
    MoCo v3~\cite{chen2021empiricalstudytrainingselfsupervised}         & ViT-B/16 & IN-1k   & 49.64\ & 70.01\ & 77.36\ & 28.2596 & 31.72\ & 57.84\ & 69.20\ & 33.0295 \\
    MAE~\cite{he2022masked}             & ViT-B/16 & IN-1k   & 43.89\ & 66.13\ & 74.56\ & 30.1382 & 22.07\ & 49.06\ & 62.63\ & 36.4724 \\
    MaskFeat~\cite{wei2022masked}        & ViT-B/16 & IN-1k   & 53.63\ & 72.23\ & 79.03\ & 27.1797 & 32.40\ & 58.92\ & 70.43\ & 32.3430 \\ 
    \bottomrule
  \end{tabularx}
\end{table*}

\begin{table*}[th!]
  \newcommand{\rowheader}{\rowcolor{Gray!20}}
  \centering
\caption{
    \textbf{Geometric Correspondence Results on the NAVI Dataset.} Evaluation of self-supervised models on geometric correspondence tasks, including 3D Recall, 2D Projection Recall, and Binned Recall, across varying viewpoint angle ranges.
}
  \label{tab:navi_geometric_correspondence}
  \setlength\tabcolsep{6pt}
  \footnotesize
  \begin{tabularx}{\linewidth}{Xll ccc rrr rrrr}
    \toprule
    & &
    & \multicolumn{3}{c}{3D Recall}  
    & \multicolumn{3}{c}{2D Recall} 
    & \multicolumn{4}{c}{Bin Recall} \\
    \cmidrule(lr){4-6} 
    \cmidrule(lr){7-9}
    \cmidrule(lr){10-13}
    \textbf{Model} & \textbf{Architecture} & \textbf{Dataset} &
    0.01m & 0.02m & 0.05m &
    5px & 25px & 50px &
    0-30° & 30-60° & 60-90° & 90-120° \\
    \midrule
\midrule \rowheader \multicolumn{13}{l}{\textit{\textbf{Self-Supervised Models (SSL)}}} \\
Jigsaw~\cite{Noroozi2016}                       & RN-50 & IN-1k & 9.13 & 19.83 & 54.94 & 0.68 & 7.45 & 16.20 & 49.15 & 26.54 & 13.06 & 7.76 \\
RotNet~\cite{Gidaris2018}                       & RN-50 & IN-1k & 11.97 & 23.21 & 55.13 & 0.92 & 9.83 & 19.38 & 58.44 & 29.82 & 14.85 & 10.26 \\
NPID~\cite{Wu2018a}                            & RN-50 & IN-1k & 18.70 & 32.11 & 63.38 & 1.57 & 15.80 & 27.47 & 69.09 & 41.51 & 22.96 & 16.62 \\
SeLa v2~\cite{caron2021unsupervisedlearningvisualfeatures} & RN-50 & IN-1k & 12.17 & 23.50 & 53.26 & 0.93 & 10.14 & 19.49 & 49.33 & 28.07 & 18.86 & 12.86 \\
NPID++~\cite{misra2019pirl}                     & RN-50 & IN-1k & 13.20 & 25.86 & 58.25 & 0.87 & 10.52 & 21.20 & 53.10 & 32.17 & 19.75 & 14.41 \\
PIRL~\cite{misra2019pirl}                       & RN-50 & IN-1k & 16.21 & 29.49 & 61.54 & 1.15 & 13.21 & 24.73 & 60.73 & 36.56 & 22.61 & 16.40 \\
ClusterFit~\cite{yan2019clusterfitimprovinggeneralizationvisual} & RN-50 & IN-1k & 10.85 & 21.49 & 56.86 & 1.86 & 9.08 & 16.94 & 43.28 & 26.57 & 17.32 & 11.61 \\
DeepCluster v2~\cite{caron2021unsupervisedlearningvisualfeatures} & RN-50 & IN-1k & 20.65 & 34.42 & 64.24 & 1.78 & 18.14 & 30.46 & 69.52 & 42.24 & 27.47 & 19.09 \\
SwAV~\cite{caron2021unsupervisedlearningvisualfeatures} & RN-50 & IN-1k & 20.20 & 33.99 & 63.20 & 1.71 & 17.60 & 29.83 & 67.11 & 42.34 & 27.23 & 18.81 \\
SimCLR~\cite{chen2020simple}                    & RN-50 & IN-1k & 16.57 & 30.68 & 61.75 & 1.09 & 13.49 & 25.80 & 60.53 & 37.77 & 23.67 & 18.27 \\
MoCo v2~\cite{chen2020improved}                 & RN-50 & IN-1k & 21.85 & 37.76 & 68.76 & 1.63 & 18.17 & 32.94 & 75.85 & 48.73 & 28.47 & 20.50 \\
SimSiam~\cite{chen2021exploring}                & RN-50 & IN-1k & 23.47 & 38.16 & 68.41 & 2.07 & 20.16 & 33.57 & 76.05 & 48.63 & 29.90 & 20.46 \\
BYOL~\cite{grill2020bootstraplatentnewapproach} & RN-50 & IN-1k & 10.81 & 21.11 & 56.81 & 2.26 & 9.02 & 16.64 & 46.24 & 26.45 & 15.82 & 10.65 \\
Barlow Twins~\cite{zbontar2021barlowtwinsselfsupervisedlearning} & RN-50 & IN-1k & 12.71 & 23.27 & 58.22 & 2.97 & 10.92 & 18.83 & 52.25 & 29.38 & 17.00 & 11.41 \\
DenseCL~\cite{wang2021dense}                    & RN-50 & IN-1k & 17.59 & 34.57 & 67.63 & 1.17 & 14.28 & 29.17 & 71.25 & 44.65 & 26.29 & 17.76 \\
DINO~\cite{caron2021emerging} & RN-50 & NAVI & 30.57 & 47.36 & 75.43 & 2.61 & 26.79 & 42.41 & 84.37 & 61.43 & 39.01 & 26.82 \\
MoCo v3~\cite{chen2021empiricalstudytrainingselfsupervised} & RN-50 & NAVI & 21.70 & 36.29 & 65.49 & 1.70 & 18.43 & 31.77 & 73.41 & 45.90 & 27.84 & 19.88 \\
DINO~\cite{caron2021emerging}                   & ViT-B/16 & IN-1k & 25.91 & 43.00 & 74.66 & 3.16 & 22.54 & 36.86 & 84.78 & 56.28 & 33.20 & 22.54 \\
iBOT~\cite{zhou2021ibot}                        & ViT-B/16 & IN-1k & 26.84 & 44.72 & 76.10 & 3.12 & 23.78 & 39.11 & 86.94 & 58.98 & 34.22 & 23.85 \\
MoCo v3~\cite{chen2021empiricalstudytrainingselfsupervised} & ViT-B/16 & IN-1k & 26.99 & 44.46 & 75.22 & 2.17 & 23.45 & 39.54 & 85.95 & 58.96 & 34.45 & 23.20 \\
MAE~\cite{he2021maskedautoencodersscalablevision} & ViT-B/16 & IN-1k & 19.21 & 32.59 & 66.82 & 2.74 & 17.16 & 27.72 & 78.17 & 46.12 & 21.16 & 11.85 \\
MaskFeat~\cite{wei2022masked}                   & ViT-B/16 & IN-1k & 22.11 & 35.16 & 65.92 & 2.08 & 19.67 & 31.37 & 86.25 & 51.50 & 22.17 & 11.00 \\
\bottomrule
\end{tabularx}
\end{table*}

\begin{table*}[th!]
  \newcommand{\rowheader}{\rowcolor{Gray!20}}
  \centering
    \caption{
        \textbf{Geometric Correspondence Results on ScanNet.} Evaluation of self-supervised models on 2D Projection Recall at varying pixel error thresholds and Binned Recall across viewpoint angle ranges.
    }
  \label{tab:scannet_geometric_correspondence}
  \setlength\tabcolsep{6pt}
  \footnotesize
  \begin{tabularx}{\linewidth}{Xll rrr rrrr}
    \toprule
    & &
    \multicolumn{3}{c}{2D Recall}  
    & \multicolumn{4}{c}{Bin Recall} \\
    \cmidrule(lr){3-5} 
    \cmidrule(lr){6-9}
    \textbf{Model} & \textbf{Architecture} &
    5px & 10px & 20px &
    0-15° & 15-30° & 30-60° & 60-180° \\
\midrule \rowheader \multicolumn{9}{l}{\textit{\textbf{Self-Supervised Models}}} \\

    Jigsaw~\cite{Noroozi2016}            & RN-50    & 9.57  & 18.18 & 27.98 & 26.11 & 19.80 & 11.16 & 4.00 \\
    RotNet~\cite{Gidaris2018}            & RN-50    & 15.74 & 25.46 & 34.15 & 37.56 & 28.52 & 13.73 & 4.29 \\
    NPID~\cite{Wu2018a}                  & RN-50    & 27.64 & 40.10 & 50.07 & 52.84 & 44.85 & 28.24 & 11.34 \\
    SeLa-v2~\cite{caron2021unsupervisedlearningvisualfeatures}            & RN-50    & 12.21 & 22.70 & 33.36 & 31.73 & 24.61 & 14.61 & 6.54 \\
    NPID++~\cite{misra2019pirl}           & RN-50    & 10.62 & 19.59 & 30.23 & 27.16 & 20.92 & 13.09 & 6.37 \\
    PIRL~\cite{misra2019pirl}             & RN-50    & 17.89 & 30.43 & 41.35 & 45.37 & 35.12 & 19.67 & 7.55 \\
    ClusterFit~\cite{yan2019clusterfitimprovinggeneralizationvisual}           & RN-50    & 26.31 & 40.92 & 51.96 & 54.96 & 46.61 & 26.67 & 10.45 \\
    DeepCluster v2~\cite{caron2021unsupervisedlearningvisualfeatures}     & RN-50    & 17.30 & 27.90 & 37.57 & 38.25 & 30.90 & 18.09 & 7.87 \\
    SwAV~\cite{caron2021unsupervisedlearningvisualfeatures}               & RN-50    & 25.41 & 38.74 & 49.86 & 52.34 & 44.20 & 27.48 & 10.23 \\
    SimCLR~\cite{chen2020simple}                    & RN-50    & 21.78 & 35.34 & 46.18 & 48.85 & 40.32 & 22.15 & 9.08 \\
    MoCo v2~\cite{chen2020improved}                 & RN-50    & 24.92 & 37.65 & 48.33 & 50.92 & 41.97 & 24.56 & 8.97 \\
    SimSiam~\cite{chen2021exploring}                & RN-50    & 18.11 & 29.83 & 40.92 & 42.58 & 33.72 & 19.04 & 7.24 \\
    BYOL~\cite{grill2020bootstraplatentnewapproach}                 & RN-50    & 15.39 & 25.41 & 34.89 & 35.88 & 26.91 & 16.96 & 6.90 \\
    Barlow Twins~\cite{zbontar2021barlowtwinsselfsupervisedlearning}      & RN-50    & 18.83 & 30.60 & 40.61 & 42.36 & 33.96 & 19.24 & 8.55 \\
    DenseCL~\cite{wang2021dense}                    & RN-50    & 17.23 & 31.17 & 44.98 & 42.41 & 34.80 & 20.36 & 8.56 \\
    DINO~\cite{caron2021emerging}                  & RN-50    & 26.63 & 40.64 & 51.49 & 54.07 & 45.63 & 27.80 & 11.19 \\
    MoCo v3~\cite{chen2021empiricalstudytrainingselfsupervised}                & RN-50    & 15.23 & 26.06 & 35.87 & 37.24 & 28.23 & 15.94 & 7.05 \\
    DINO~\cite{caron2021emerging}                  & ViT-B/16 & 24.38 & 34.22 & 45.47 & 46.56 & 36.72 & 23.74 & 11.12 \\
    iBOT~\cite{zhou2021ibot}                        & ViT-B/16 & 20.04 & 29.45 & 41.07 & 41.13 & 30.95 & 20.00 & 9.47 \\
    MoCo v3~\cite{chen2021empiricalstudytrainingselfsupervised}                & ViT-B/16 & 25.03 & 39.31 & 51.00 & 53.18 & 42.87 & 27.05 & 11.95 \\
    MAE~\cite{he2022masked}                            & ViT-B/16 & 6.64  & 10.31 & 18.42 & 15.64 & 9.81  & 6.63  & 3.81 \\
    MaskFeat~\cite{wei2022masked}                   & ViT-B/16 & 27.94 & 40.87 & 50.49 & 56.51 & 47.65 & 24.41 & 6.90 \\
    \bottomrule
  \end{tabularx}
\end{table*}

\begin{table*}[th!]
  \newcommand{\rowheader}{\rowcolor{Gray!20}}
  \centering
    \caption{
        \textbf{Image Retrieval Results on the NIGHTS Dataset.} Retrieval performance is evaluated using Accuracy, F1-score, Precision, and Recall. Results are reported for self-supervised models using ResNet-50 and ViT-B/16 backbones, highlighting their capability to retrieve similar images based on mid-level features.
    }
    \label{tab:image_retrieval_nights}
  \setlength\tabcolsep{6pt}
  \footnotesize
  \begin{tabularx}{\linewidth}{Xll ccc ccc}
    \toprule
    \textbf{Model} & \textbf{Backbone} & \textbf{Dataset} &
    Accuracy & F1-Score & Precision & Recall \\
    \midrule \rowheader \multicolumn{7}{l}{\textit{\textbf{Self-Supervised Models (SSL)}}} \\
    Jigsaw~\cite{Noroozi2016}               & RN-50      & NIGHTS & 71.22 & 70.69 & 70.73 & 70.65 \\
    RotNet~\cite{Gidaris2018}             & RN-50      & NIGHTS & 75.33 & 75.14 & 74.40 & 75.89 \\
    NPID~\cite{Wu2018a}                 & RN-50      & NIGHTS & 81.41 & 81.16 & 80.84 & 81.47 \\
    SeLa-v2~\cite{caron2021unsupervisedlearningvisualfeatures}       & RN-50      & NIGHTS & 81.41 & 81.16 & 80.84 & 81.47 \\
    NPID++~\cite{misra2019pirl}         & RN-50      & NIGHTS & 83.06 & 82.63 & 83.24 & 82.03 \\
    PIRL~\cite{misra2019pirl}            & RN-50      & NIGHTS & 83.77 & 83.56 & 83.19 & 83.93 \\
    ClusterFit~\cite{yan2019clusterfitimprovinggeneralizationvisual} & RN-50      & NIGHTS & 81.58 & 81.42 & 80.70 & 82.14 \\
    DeepCluster-v2~\cite{caron2021unsupervisedlearningvisualfeatures} & RN-50      & NIGHTS & 85.25 & 84.93 & 85.26 & 84.60 \\
    SwAV~\cite{caron2021unsupervisedlearningvisualfeatures}          & RN-50      & NIGHTS & 84.65 & 84.36 & 84.45 & 84.26 \\
    SimCLR~\cite{chen2020simple}         & RN-50      & NIGHTS & 83.55 & 83.26 & 83.26 & 83.26 \\
    MoCo v2~\cite{chen2020improved}      & RN-50      & NIGHTS & 84.43 & 84.22 & 83.85 & 84.60 \\
    SimSiam~\cite{chen2021exploring}     & RN-50      & NIGHTS & 85.86 & 85.78 & 84.75 & 86.83 \\
    BYOL~\cite{grill2020bootstraplatentnewapproach}    & RN-50      & NIGHTS & 85.86 & 85.75 & 84.90 & 86.61 \\
    Barlow Twins~\cite{zbontar2021barlowtwinsselfsupervisedlearning} & RN-50      & NIGHTS & 83.11 & 82.70 & 83.26 & 82.14 \\
    DenseCL~\cite{wang2021dense}         & RN-50      & NIGHTS & 82.73 & 82.53 & 82.03 & 83.04 \\
    DINO~\cite{caron2021emerging}         & RN-50      & NIGHTS & 83.83 & 83.31 & 84.50 & 82.14 \\
    MoCo v3~\cite{chen2021empiricalstudytrainingselfsupervised}      & RN-50      & NIGHTS & 84.70 & 84.37 & 84.70 & 84.04 \\
    DINO~\cite{caron2021emerging}         & ViT-B/16   & NIGHTS & 89.20 & 88.98 & 89.23 & 88.73 \\
    iBOT~\cite{zhou2021ibot}            & ViT-B/16   & NIGHTS & 89.36 & 89.27 & 88.49 & 90.07 \\
    MoCo v3~\cite{chen2021empiricalstudytrainingselfsupervised}      & ViT-B/16   & NIGHTS & 87.17 & 86.90 & 87.19 & 86.61 \\
    MAE~\cite{he2021maskedautoencodersscalablevision}  & ViT-B/16   & NIGHTS & 83.39 & 82.91 & 83.81 & 82.03 \\
    MaskFeat~\cite{wei2022masked}        & ViT-B/16   & NIGHTS & 76.10 & 75.70 & 75.61 & 75.78 \\
    \bottomrule
  \end{tabularx}
\end{table*}

\end{document}